%% file: main.tex
\documentclass[10pt,twocolumn,letterpaper]{article}

\usepackage{cvpr}              

\input{preamble}

\definecolor{cvprblue}{rgb}{0.21,0.49,0.74}
\usepackage[pagebackref,breaklinks,colorlinks,allcolors=cvprblue]{hyperref}

\def\paperID{2795} 
\def\confName{CVPR}
\def\confYear{2026}

\title{SeeThrough3D: Occlusion Aware 3D Control in Text-to-Image Generation}

\author{
  \begin{tabular}[t]{c}
    Vaibhav Agrawal$^{1}$ \quad
    Rishubh Parihar$^{2}$ \quad
    Pradhaan S Bhat$^{2}$ \quad \\ 
    Ravi Kiran Sarvadevabhatla$^{1}$\footnotemark[2] \quad 
    R. Venkatesh Babu$^{2}$\footnotemark[2] \\ 
    \end{tabular}%
  \quad
  \and
  \begin{tabular}[t]{c} 
    $^1$IIIT Hyderabad \quad 
    $^2$IISc Bengaluru \\
  \end{tabular}%
}

\begin{document}

\twocolumn[{
    \vspace{-16mm}
    \begin{@twocolumnfalse}  
        \maketitle
        \thispagestyle{plain}  
        \begin{center}
            \includegraphics[width=0.85\textwidth]{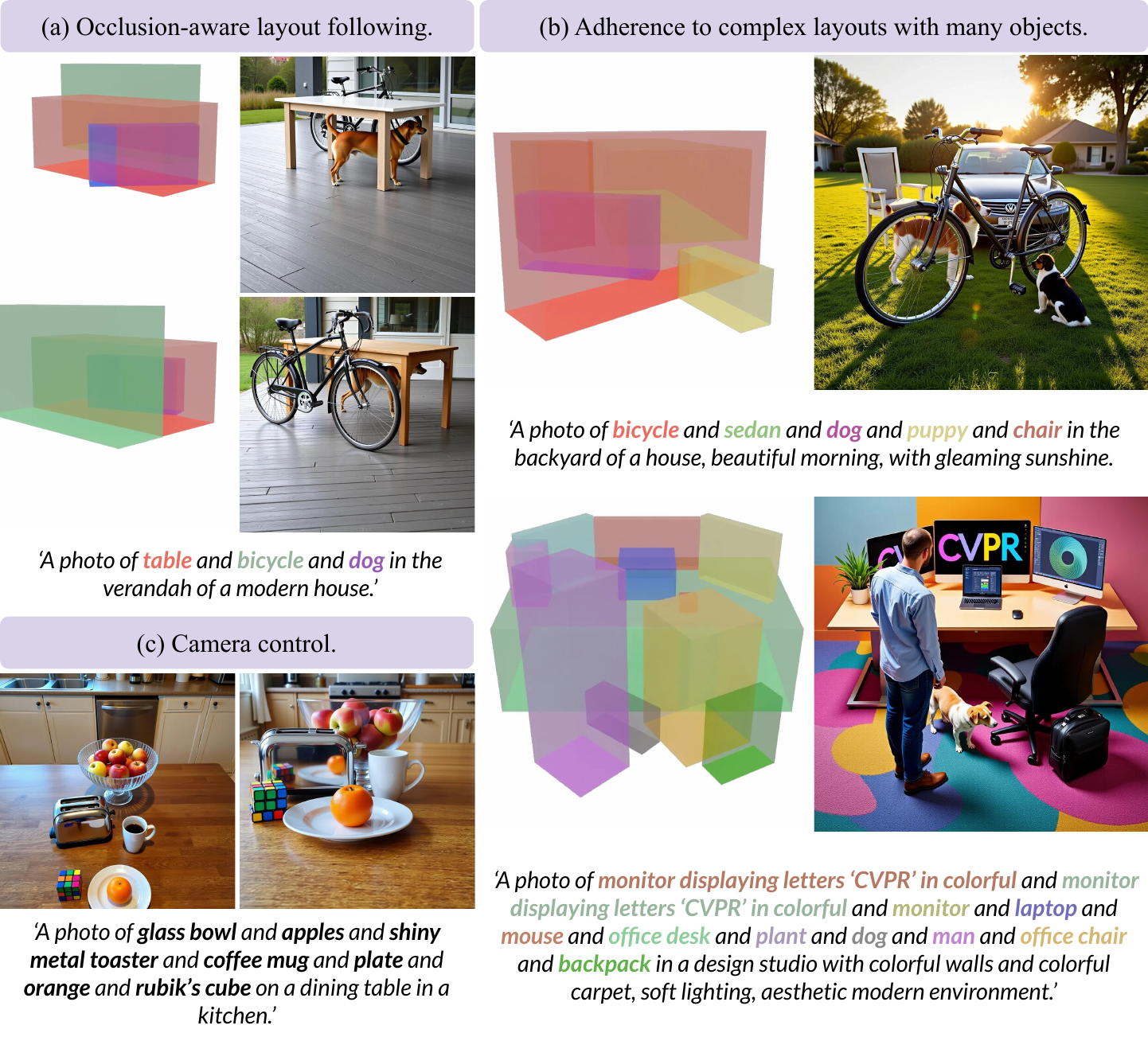}
            \captionsetup{type=figure}
            \captionof{figure}{We propose \textbf{SeeThrough3D}, a method for occlusion aware 3D scene control in text-to-image generation. Our method enables (a) occlusion-aware 3D object placement in generated images, and (b) adheres well to complex layouts featuring many objects. Additionally, our method allows for (c) control over the camera viewpoint in the generated image.}
            \label{fig:teaser}
        \end{center}
    \end{@twocolumnfalse}
}]

\renewcommand{\thefootnote}{\fnsymbol{footnote}} 
\footnotetext[2]{Equal advising}

\input{sec/0_abstract}    
\input{sec/1_intro}

\input{sec/2_related_work}

\input{sec/3_method}

\input{sec/4_experiments}
\vspace{-0.5mm}
\input{sec/5_conclusion}
{
    \small
    \bibliographystyle{ieeenat_fullname}
    \bibliography{main}
}

\input{sec/X_suppl}


\end{document}

%% file: preamble.tex
\newcommand{\VA}[1]{}

\usepackage[table,xcdraw]{xcolor}
\usepackage{soul}
\usepackage[normalem]{ulem}
\usepackage[hang,flushmargin]{footmisc}
\usepackage[most]{tcolorbox}
\usepackage{circledsteps}

\definecolor{lightgray}{RGB}{226,226,226}  
\definecolor{darkpink}{RGB}{102,0,102}  
\definecolor{lightpink}{RGB}{255,210,239} 
\definecolor{darkbrown}{RGB}{102,51,0} 
\definecolor{darkblue}{RGB}{0,0,153}  
\definecolor{palegreen}{RGB}{0,108,108}  
\definecolor{mypurple}{RGB}{67,67,67}
\definecolor{myyellow}{RGB}{212, 167, 16}
\definecolor{myblue}{RGB}{108, 142, 191} 
\definecolor{mygreen}{RGB}{96, 169, 23} 
\definecolor{Mgreen}{RGB}{0, 102, 51} 
\definecolor{highlightyellow}{RGB}{247, 210, 99}

\pgfkeys{/csteps/inner color=white}
\pgfkeys{/csteps/outer color=darkbrown}
\pgfkeys{/csteps/fill color=darkbrown}

\DeclareRobustCommand{\CircledColored}[2]{%
\begingroup
\pgfkeys{/csteps/.cd,
inner color=white,
outer color=#1,
fill color=#1}%
\Circled{#2}%
\endgroup
}

\DeclareRobustCommand{\CircledMask}[1]{\CircledColored{Mgreen}{#1}}

\usepackage{placeins} 

%% file: sec/0_abstract.tex
\begin{abstract}

We identify occlusion reasoning as a fundamental yet overlooked aspect for 3D layout–conditioned generation. It is essential for synthesizing partially occluded objects with depth-consistent geometry and scale. While existing methods can generate realistic scenes that follow input layouts, they often fail to model precise inter-object occlusions. We propose \textbf{SeeThrough3D}, a model for 3D layout conditioned generation that explicitly models occlusions. We introduce an occlusion-aware 3D scene representation (OSCR), where objects are depicted as translucent 3D boxes placed within a virtual environment and rendered from desired camera viewpoint. The transparency encodes hidden object regions, enabling the model to reason about occlusions, while the rendered viewpoint provides explicit camera control during generation. We condition a pretrained flow based text-to-image image generation model by introducing a set of visual tokens derived from our rendered 3D representation. Furthermore, we apply masked self-attention to accurately bind each object bounding box to its corresponding textual description, enabling accurate generation of multiple objects without object attribute mixing. To train the model, we construct a synthetic dataset with diverse multi-object scenes with strong inter-object occlusions. SeeThrough3D generalizes effectively to unseen object categories and enables precise 3D layout control with realistic occlusions and consistent camera control. Project page: \url{https://seethrough3d.github.io}




\end{abstract}

%% file: sec/1_intro.tex
\vspace{-8mm}
\textbf{}\section{Introduction}
\label{sec:intro}



Recent work has introduced various forms of controllability in text-to-image generation, but most methods remain limited to 2D spatial controls, such as bounding boxes or segmentation
maps~\cite{controlnet,mo2024freecontrol,tan2024ominicontrol,tan2025ominicontrol2,zhang2025easycontrol,fu2025univg,parihar2024text2place}. While effective for coarse control over the scene content, they offer limited control over inherently 3D scene properties, including object arrangement and camera viewpoint. Yet many practical content-creation domains such as design, gaming, and architectural visualization require precise 3D layout control, where object size, orientation, and placement must be explicitly specified. Critically, a truly 3D-aware generative model must also reason about occlusions, generate partially hidden objects with depth-consistent scale and perspective; a fundamental capability that 2D controls cannot provide.

Despite being fundamental to accurate 3D-aware generation, occlusion has been largely overlooked in recent 3D layout based methods. Existing approaches condition the generative model on depth maps derived from 3D bounding-box layouts~\cite{loosecontrol,wang2025cinemaster} or on explicit 3D attributes such as object or camera poses~\cite{parihar2025compass,cont-words,qinscenedesigner2025,view-neti,min2025origen}. These methods succeed in generating simple scenes with few objects and minimal occlusion, but fail to model significant inter-object occlusions in multi-object layouts (\cref{fig:compare_representations}(a)). A related direction represents scenes as a stack of 2D object layers~\cite{zhan2025larender, liang2025vodiff} to approximate occlusion, but this collapses the inherently 3D structure of the scene into flat planes (\cref{fig:compare_representations}(c)), leading to generating object occlusion that violate true 3D geometry and perspective.




In this paper, we propose \textit{SeeThrough3D} - an image generation model that takes 3D layout and text prompt as input and generates scenes with 3D consistent occlusions (\cref{fig:teaser}). We introduce an efficient and expressive 3D scene representation, termed \textbf{O}cclusion-Aware 3D \textbf{Sc}ene \textbf{R}epresentation (\textbf{OSCR}), which jointly encodes object arrangements and camera viewpoint (\cref{fig:oscr}). In OSCR, each object is modeled as a translucent 3D bounding box, where transparency reveals occluded regions, enabling explicit reasoning about inter-object occlusions. Faces of each box are further color-coded according to a predefined mapping to capture 3D object orientation. The final OSCR representation is obtained by rendering this layout from a specified camera viewpoint.  



We build on FLUX~\cite{labs2025flux} image generator, conditioning it on our OSCR scene representation. Following the success of recent works~\cite{tan2024ominicontrol, labs2025flux} on controlling the diffusion transformer (DiT)~\cite{peebles2023scalable,esser2024scaling} using condition image tokens, we condition the model with tokens derived from our rendered scene representation. However, spatial conditioning alone fails to associate textual object descriptions with their corresponding box regions. To address this, we apply attention masking to bind each object to its corresponding box, ensuring accurate bounding box adherence for individual objects. Further, we extend this framework to allow 3D control of personalized objects, by conditioning on an image of the object, and binding its appearance to specific box in the OSCR representation.  



To train SeeThrough3D, we create a synthetic dataset of scenes by placing diverse 3D assets in a virtual environment~\cite{blender} and rendering scenes from multiple camera views. Object placement and camera parameters are controlled to induce strong inter-object occlusions in the rendered images. Despite being trained on synthetic data, SeeThrough3D generalizes well to unseen objects, backgrounds and complex scene layouts (see~\cref{fig:teaser}), evaluated qualitatively and through metrics, as well as a user study.

%% file: sec/2_related_work.tex
\vspace{-1mm} 
\section{Related work}
\label{sec:related_work}

\textbf{3D control in text-to-image generation:} Previous works on 3D control in image generation trains specialized generative models conditioned on various 3D representations~\cite{wang2023blobgan,nguyen2019hologan, niemeyer2021giraffe, xue2022giraffehd, blendnerf, bautista2022gaudi, Kathare_2025_WACV}. Interestingly, recent works have shown that there is inherent 3D understanding in large text-to-image diffusion models~\cite{diffusion3d-understand1,diffusion3d-understand2,diffusion3d-understand3}. Several works leverage this insight for enabling precise 3D aware control in generated images~\cite{view-neti,dhiman2024reflecting,sajnani2024geodiffuser,kumari2024customizing,bernal2025precisecam,cheng20253d,escontrela2025neural,Higgins_2025_ICCV,view-neti,zero-123}. One line of works enable 3D aware editing~\cite{wang2024diffusioncritc,diffusion-handles,sajnani2024geodiffuser} using scene depth as additional input, but they are limited to manipulation of a single object at a time. Further, a recent work~\cite{mdf} decomposes a scene into depth-based layers, enabling depth-aware editing and scene composition. Others train implicit 3D representations such as radiance fields~\cite{patashnik2024consolidating,cd-360,yenphraphai2024image} or 3D Gaussian splats~\cite{zhang20243ditscene,chen2023gaussianeditor,luo20243d,wen2025intergsedit,kohdiffusion} in diffusion feature space to enable 3D aware image editing. 

\noindent
\textbf{3D layout conditioned generation:} Apart from editing, controlling the 3D layout of a scene during generation is an active research area. A recent work for layout-conditioned generation, LooseControl~\cite{loosecontrol} conditions a text-to-image model using depth maps of 3D bounding boxes; however it fails to generate complex scenes with diverse objects. A follow-up work, Build-A-Scene~\cite{eldesokey2024build} generates the scene using multiple generation-inversion cycles, each iteration adding a new object. However, this leads to inversion artifacts and incoherence in generated images. Another set of works provide partial control over individual 3D properties, such as object orientation~\cite{min2025origen,parihar2025compass,cont-words}, but they are limited in their extent to precisely control object placement or camera viewpoint. Another promising direction for 3D layout control is to represent the object bounding box as a set and condition the generative model using a learnable adapter~\cite{neural-assets,maillard2025laconic,parihar2025monoplace3d}. However, they are limited to a single data domain, e.g. road scenes or indoor scenes, and are less effective than spatial conditioning approaches~\cite{qinscenedesigner2025}.

\noindent 
\textbf{Occlusion awareness:} Inter-object occlusions present a significant challenge in perception~\cite{kortylewski2020combining,kassaw2025deep,fawzi2016measuring,mallick2025d,zhan2024amodal,tai2025segment,Li_2025_ICCV} and generation~\cite{parihar2025compass,zhan2025larender,liang2025vodiff,wu2025amodal3r,liu2024object} tasks. Occlusions are particularly important for 3D aware image generation. However, it has received little attention in existing works~\cite{loosecontrol,eldesokey2024build}. Some works model occlusions by decomposing images into flat 2D object layers~\cite{zhan2025larender,liang2025vodiff, damaraju2025cobl}, but they lack 3D awareness, resulting in geometrically inconsistent occlusions. To bridge the gap in existing works, we propose SeeThrough3D, a model that enables generalized occlusion-aware 3D layout control.

%% file: sec/3_method.tex
\section{Method}
\vspace{-1mm}

\begin{figure}[t]
  \centering
   \includegraphics[width=1.0\linewidth]{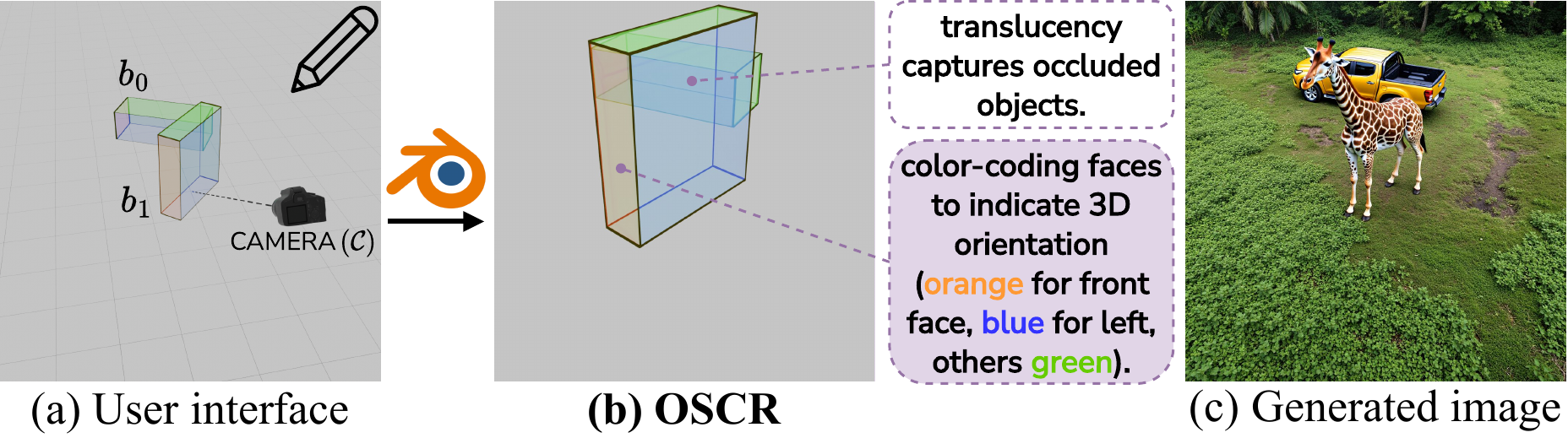}
   \vspace{-4mm}
   \caption{\textbf{OSCR:} We propose \underline{O}cclusion-Aware \underline{Sc}ene \underline{R}epresentation (OSCR) for 3D layout control in text-to-image generation. OSCR describes objects as translucent 3D boxes, which exposes occluded regions, enabling the generative model to reason about occlusions. Further, each box face is color-coded with a mapping to encode its 3D orientation. (a) A user specifies the object bounding boxes ($b_0$ and $b_1$) and sets desired viewpoint $\mathcal{C}$ in an interactive graphic environment. (b) These boxes are rendered to obtain our OSCR representation, (c) which is used to condition the generation for occlusion aware 3D control.} 
   \label{fig:oscr}
   \vspace{-5mm}
\end{figure}

Our goal is to generate an image conditioned on a text prompt and a scene layout consisting of 3D bounding boxes. We build on a pretrained text-to-image flow model~\cite{labs2025flux} and condition on the proposed Occlusion-Aware 3D Scene Representation (OSCR) (see~\cref{fig:oscr}). 




\subsection{OSCR}
\begin{figure}[!h]
  \centering
   \includegraphics[width=1.0\linewidth]{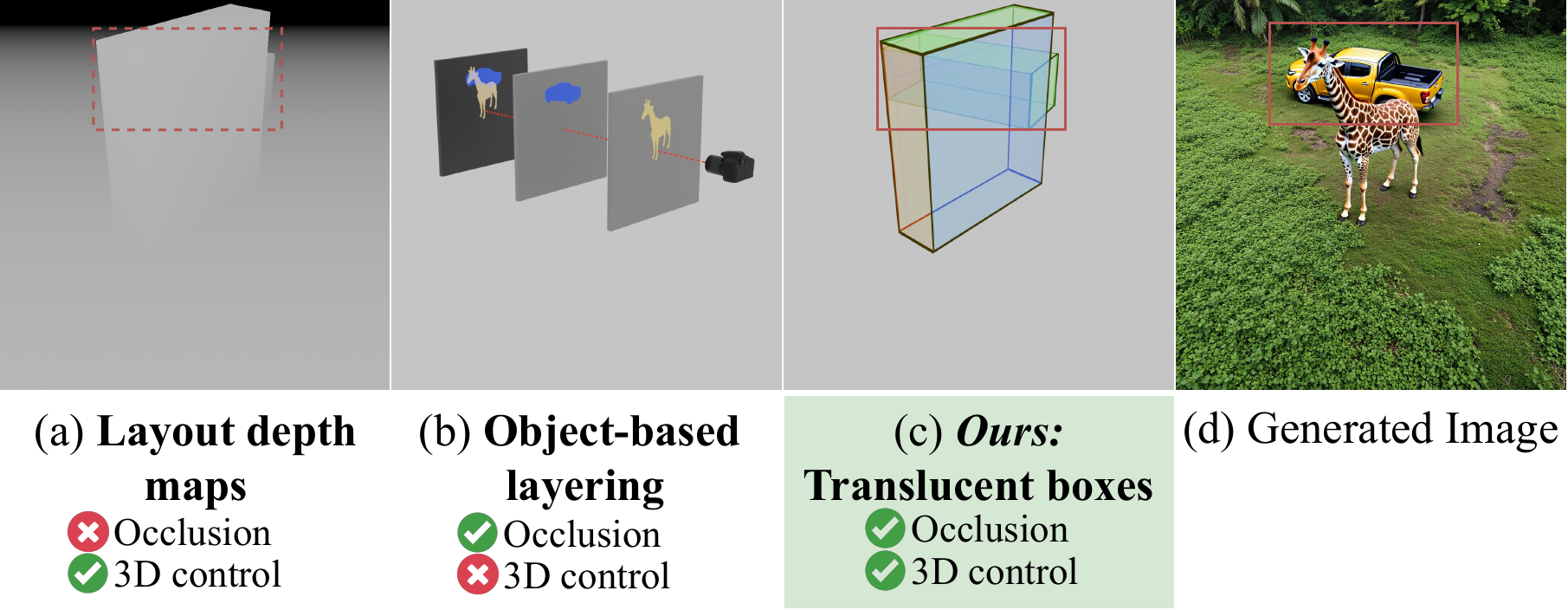}
   \vspace{-2mm}
   \caption{\textbf{Towards occlusion aware 3D scene layouts:} existing methods represent scenes as (a) 3D layout depth maps~\cite{loosecontrol,eldesokey2024build,wang2025cinemaster}, which fail to represent occluded objects (see dashed red box), or (b) object layers~\cite{zhan2025larender,liang2025vodiff}, which are not 3D aware, hence fail to capture camera viewpoint and perspective. (c) Therefore, we propose OSCR, where objects are described using translucent 3D bounding boxes. The transparency exposes occluded regions (red box), providing cues for occlusion reasoning, while enabling 3D layout control.}  
   \label{fig:compare_representations}
   \vspace{-4mm}
\end{figure}

\noindent 
Existing methods for 3D layout–conditioned generation represent scene layouts either by computing depth maps of 3D bounding boxes (see~\cref{fig:compare_representations}(a)) or by simplifying the scene into a finite set of 2D object layers (\cref{fig:compare_representations}(b)). These representations, however, fail to capture true 3D structure of the scene, resulting in inaccurate occlusion modeling and limited orientation control. To overcome this, we design OSCR, an efficient yet effective representation that encodes 3D layouts in an occlusion-aware manner.

Our input is a set of 3D bounding boxes ${b_i}$, each representing an object, arranged in a 3D virtual environment (see~\cref{fig:oscr}(a)). To encode object orientation, we define a canonical color mapping across box faces, where each face is assigned a predefined color (see~\cref{fig:oscr}(b)). This mapping provides an explicit and interpretable encoding of 3D orientation directly in image space. To make the representation aware of spatial ordering and occlusions, we render the boxes as translucent, allowing occluded objects to remain partially visible. This simple yet expressive design compactly captures both orientation and occlusion cues (see~\cref{fig:oscr}(b)). Notably, occlusion may alter the apparent colors of some faces, causing them to deviate from the predefined mapping. However, the relative color differences between faces remain discernible, preserving reliable orientation cues. Finally, we render the composed scene from a specified camera view $\mathcal{C}$ using Blender~\cite{blender}. The rendered image inherently embeds camera pose information, enabling precise viewpoint control in generation. The rendered image $r$ is used as `OSCR condition' to the generative model (see~\cref{fig:method_overview}).

\subsection{SeeThrough3D}
\begin{figure}[!t]
  \centering
   \includegraphics[width=0.9\linewidth]{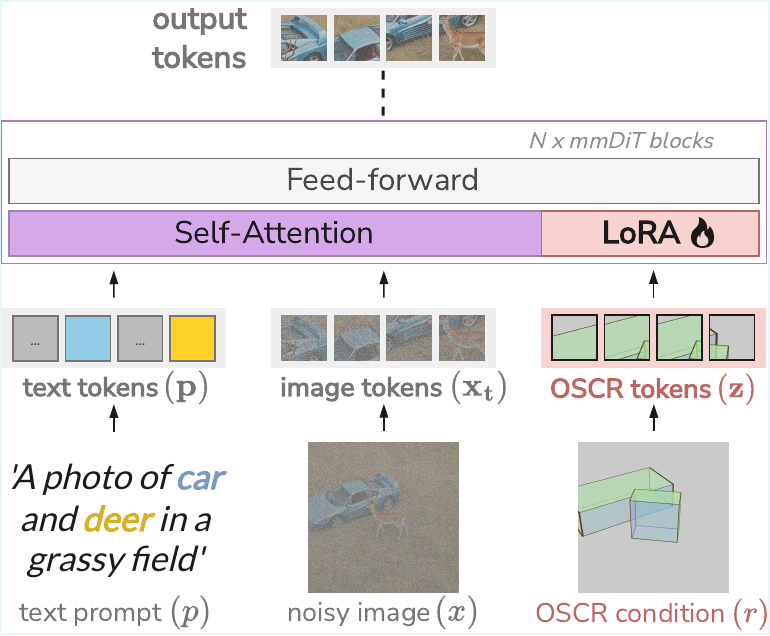}
    \vspace{-2mm}
   \caption{\textbf{SeeThrough3D:} We encode the rendered OSCR condition map r using the VAE to obtain OSCR tokens. These are concatenated with text prompt tokens $\mathbf{p}$ and noisy image tokens $\mathbf{x}_t$. The concatenated result is passed through the DiT based text-to-image model where they are jointly processed using self attention modules. We inject LoRA~\cite{hu2021lora} onto the attention projections corresponding to OSCR tokens; this enables control while preserving prior of the base model~\cite{zhang2025easycontrol,tan2024ominicontrol,tan2025ominicontrol2}.} 
   \label{fig:method_overview}\vspace{-5mm} 
   
\end{figure}

\begin{figure}[!t]
  \centering
   \includegraphics[width=1.0\linewidth]{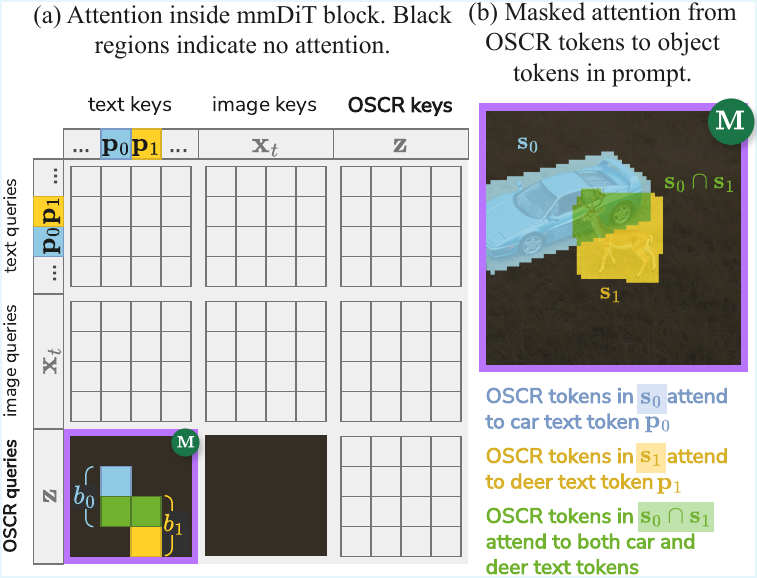}
   \vspace{-6mm}
   \caption{(a) Inside the mmDiT block, text tokens $\mathbf{p}$, image tokens $\mathbf{x}_t$ and OSCR tokens $\mathbf{z}$ are jointly processed using self attention, conditioning the generation on our OSCR representation. To bind objects to corresponding boxes, we mask the attention to enable OSCR tokens within each box $\{ b_i \}$ to attend to corresponding object tokens $\{ \mathbf{p}_i \}$ using a mask~\CircledMask{$\mathbf{M}$} (b) For this, we require spatial extent for each object box $b_i$, which we obtain we use its amodal segmentation mask $\mathbf{s}_i$. When multiple boxes overlap, their region of intersection (green) attends to multiple objects.} 
   \label{fig:masked_attention}
   \vspace{-4mm}
\end{figure}

We build on FLUX~\cite{labs2025flux}, a DiT-based text-to-image model. FLUX comprises a series of multimodal DiT blocks that jointly process text and image tokens through self-attention and feed-forward layers (see Fig.~\ref{fig:method_overview}). This architecture facilitates rich information exchange between text and image tokens, resulting in strong image-text alignment during generation. Further, this design naturally supports an effective way to condition the model on a new modality by adding condition tokens~\cite{tan2024ominicontrol,tan2025ominicontrol2,zhang2025easycontrol}. Leveraging this, we condition the model on the rendered OSCR layout representation $r$ (see~\cref{fig:method_overview}). Specifically, we first encode $r$ using the VAE to obtain OSCR tokens $\mathbf{z}$, which are concatenated with text prompt tokens $\mathbf{p}$ and the noisy image tokens $\mathbf{x_t}$. The OSCR tokens $\mathbf{z}$ are assigned the same positional encodings as the noisy image tokens $\mathbf{x}_t$, establishing spatial correspondence between them. The combined token sequence is then processed by mmDiT blocks. To adapt the model to OSCR condition while preserving its text-to-image prior, we train a LoRA~\cite{hu2021lora} only on the projection matrices associated with the newly added tokens (see~\cref{fig:method_overview}). In line with recent work~\cite{zhang2025easycontrol}, we also block attention from OSCR tokens $\mathbf{z}$ to the image tokens $\mathbf{x}_t$ (see~\cref{fig:masked_attention}).

\subsection{Object binding with attention masking}

While the conditioning mechanism described above ensures spatial alignment with the given layout, it does not explicitly associate 3D bounding boxes with their corresponding object identities. This ambiguity arises because OSCR encodes geometric arrangements of objects but lacks semantic information about them, which can lead to mismatched object placements during generation. A straightforward solution would be to encode object classes as colors within the boxes, similar to semantic segmentation. However, this approach constrains the model to a fixed set of predefined categories and limits generalization. Instead, we utilize the attention mechanism to enrich OSCR tokens with corresponding object semantics. Specifically, we mask the attention so that OSCR tokens $\mathbf{z}$ within each bounding box only attend to corresponding object noun tokens $\mathbf{p}_i$ in the text prompt,  (see~\cref{fig:masked_attention}(a)~\CircledMask{$\mathbf{M}$}), thus enriching the spatial OSCR tokens with corresponding object semantics. For this, we require the spatial extents for each box $b_i$, which we obtain using its rendered segmentation mask ,  (see~\cref{fig:masked_attention}(b)) using Blender.

\begin{figure}[!t]
  \centering
   \includegraphics[width=1.0\linewidth]{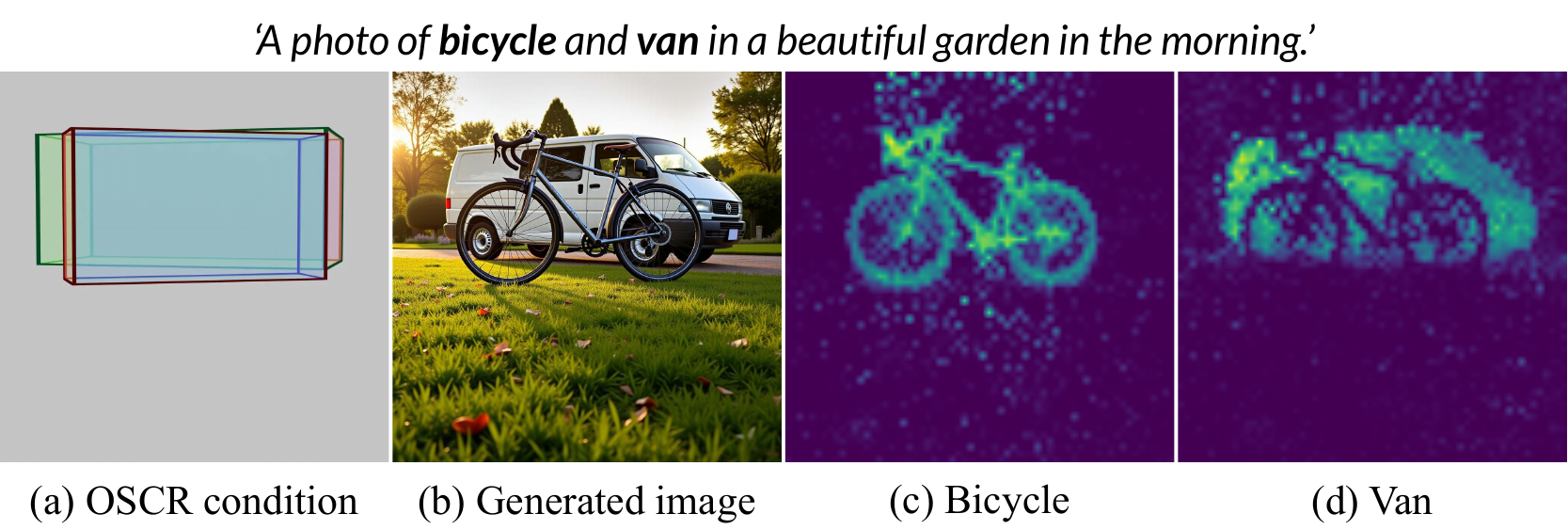}
    \vspace{-6mm}
   \caption{\textbf{Visualizing object disentanglement in latent space:} Given a layout with heavy occlusion like (a), our model's outputs show precise occlusion boundaries (b). To understand this, we visualize attention from image-tokens to object tokens in prompt (bicycle and van). Interestingly, the attention maps themselves reveal occlusion boundaries: inside the empty regions of the bicycle structure, attention on the van remains visible, accurately reflecting its presence behind the bicycle. This suggests that object-specific features remain distinct in the model’s latent space, indicating strong priors for occlusion reasoning.} 
   \label{fig:attn_viz}
   \vspace{-5mm}
\end{figure}

\vspace{1mm}
\noindent
\textbf{Handling overlapping objects:} A challenging case for the proposed object binding arises when the rendered regions of two boxes significantly overlap. In this scenario, the OSCR tokens in the intersection region attend to multiple object tokens (see~\cref{fig:masked_attention}(b)). At first glance, it appears that attending to multiple objects would lead to semantic blending or visual artifacts at object boundaries. To investigate this, we condition our model on a complex layout with heavy occlusion (see~\cref{fig:attn_viz}(a)), and observe that the output contains precise occlusion boundaries (see~\cref{fig:attn_viz}(b)). To understand this further, we visualize attention from image tokens $\mathbf{x}_t$ to object tokens $\{ \mathbf{p}_i \}$ in~\cref{fig:attn_viz}(c,d) Interestingly, the attention maps themselves reveal occlusion boundaries: inside the empty regions of the bicycle structure, attention on the van remains visible, accurately reflecting its presence behind the bicycle. This indicates that object-specific features remain distinct in the model’s latent space, and that the text-to-image model encodes necessary priors for occlusion reasoning. Our OSCR representation (see~\cref{fig:attn_viz}(a)) leverages these priors for precise control over scene layout, in an occlusion-aware manner. Further analysis of attention is provided in appendix~\cref{sec:attention_analysis}.


\subsection{Personalization} 
\vspace{-1mm}
The proposed method naturally supports layout-conditioned generation with personalized objects. Given a reference object image $v$, a text prompt $p$, and OSCR layout $r$, the goal is to generate the object adhering to a specific 3D bounding box $b_i$ in the layout $r$. We first encode object appearance by passing the reference image $v$ through the VAE encoder, resulting in `appearance tokens' $\mathbf{v}$. These are concatenated with text tokens $\mathbf{p}$, target image tokens $\mathbf{x}_t$, and OSCR tokens $\mathbf{z}$ before passing through the mmDiT blocks. To bind the object’s appearance to its corresponding 3D box $b_i$, we re-use the attention masking strategy described above. Specifically, we enable OSCR tokens inside the segmentation mask $\mathbf{s}_i$ to attend to appearance tokens $\mathbf{v}$. This enables layout-aware generation of personal objects, and can be extended to multiple objects by adding separate appearance token sets for each reference image (see~\cref{fig:personalization}). 



\vspace{-1mm}
\subsection{Dataset}
\label{subsec:dataset}
\vspace{-1mm}

To adapt the model to OSCR representation, we require a dataset of paired images and 3D bounding boxes. While existing 3D object detection datasets~\cite{Cordts_2016_CVPR,song2015sun} could be used, they are often domain specific, lack occlusion scenarios, have minimal viewpoint variation and contain marginal errors in 3D annotations, making them unsuitable for our purposes. Therefore, we create a synthetic dataset using Blender~\cite{blender}; where we procedurally place 3D assets in controlled configurations on the floor (x-y plane). Next, we render the paired ground truth image and OSCR representation from diverse camera viewpoints. We discard trivial scenes with minimal object overlap or very low visibility of any object, as we find such filtering crucial for maintaining occlusion consistency in the generated results (see~\cref{subsec:ablations}).  

\noindent
\textbf{Augmentations:} Training solely on rendered images risks overfitting to synthetic backgrounds~\cite{parihar2025compass,cont-words}, due to limited realism and lack of diversity in object appearance and backgrounds. Since creating highly varied 3D scenes is an expensive process, we adopt a scalable alternative. We generate realistic augmentations for the rendered images, that follow the same layout but are rich in terms of appearance diversity. For each rendered image, we extract its depth and feed it through a depth-to-image generation pipeline (FLUX.1-Depth-dev)~\cite{labs2025flux} to synthesize realistic images that preserve the same spatial layout. Although this pipeline produces high-quality results, it occasionally misaligns objects with their intended depth regions, causing incorrect placements. We mitigate this by applying object-level CLIP-based filtering~\cite{radford2021learningtransferablevisualmodels} to retain only those augmentations that adhere to the original layout.
Our final dataset comprises $25K$ rendered images and $25K$ augmentations. Further details about dataset pipeline and dataset statistics are provided in appendix~\cref{sec:appendix_dataset}.

\begin{figure}[!t]
  \centering
   \includegraphics[width=1.0\linewidth]{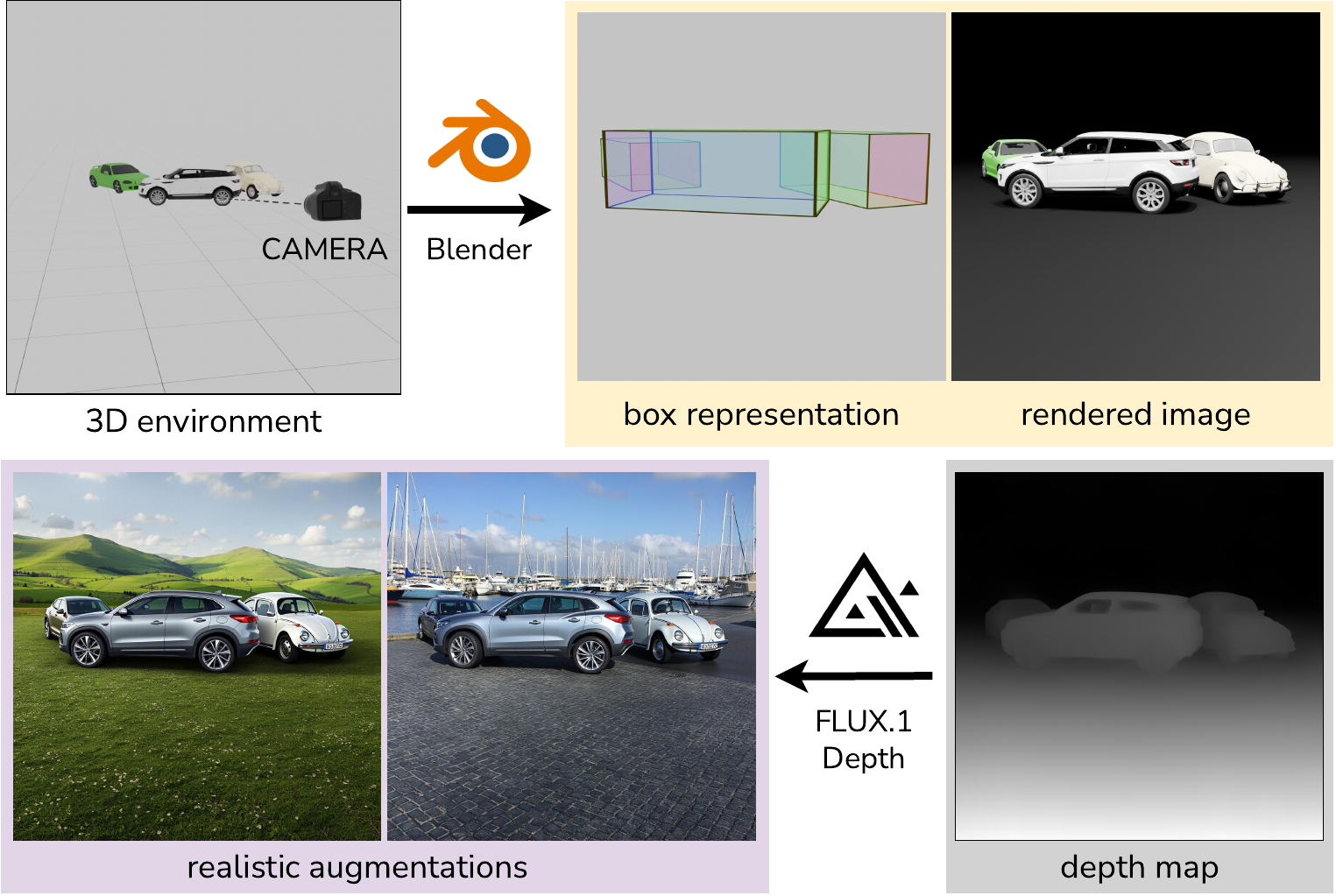}
   \vspace{-4mm}
   \caption{\textbf{Dataset creation:} We place 3D assets in controlled configurations in Blender~\cite{blender}. Object placements and camera viewpoint are controlled to ensure strong occlusions, while ensuring adequate visibility for each object. To generate realistic augmentations, we estimate image depth, and pass it through a depth-to-image model~\cite{labs2025flux} with diverse background prompts.} 
   \label{fig:dataset}
   \vspace{-6mm}
\end{figure}


\begin{figure*}[!]
  \centering
   \includegraphics[width=1.0\linewidth]{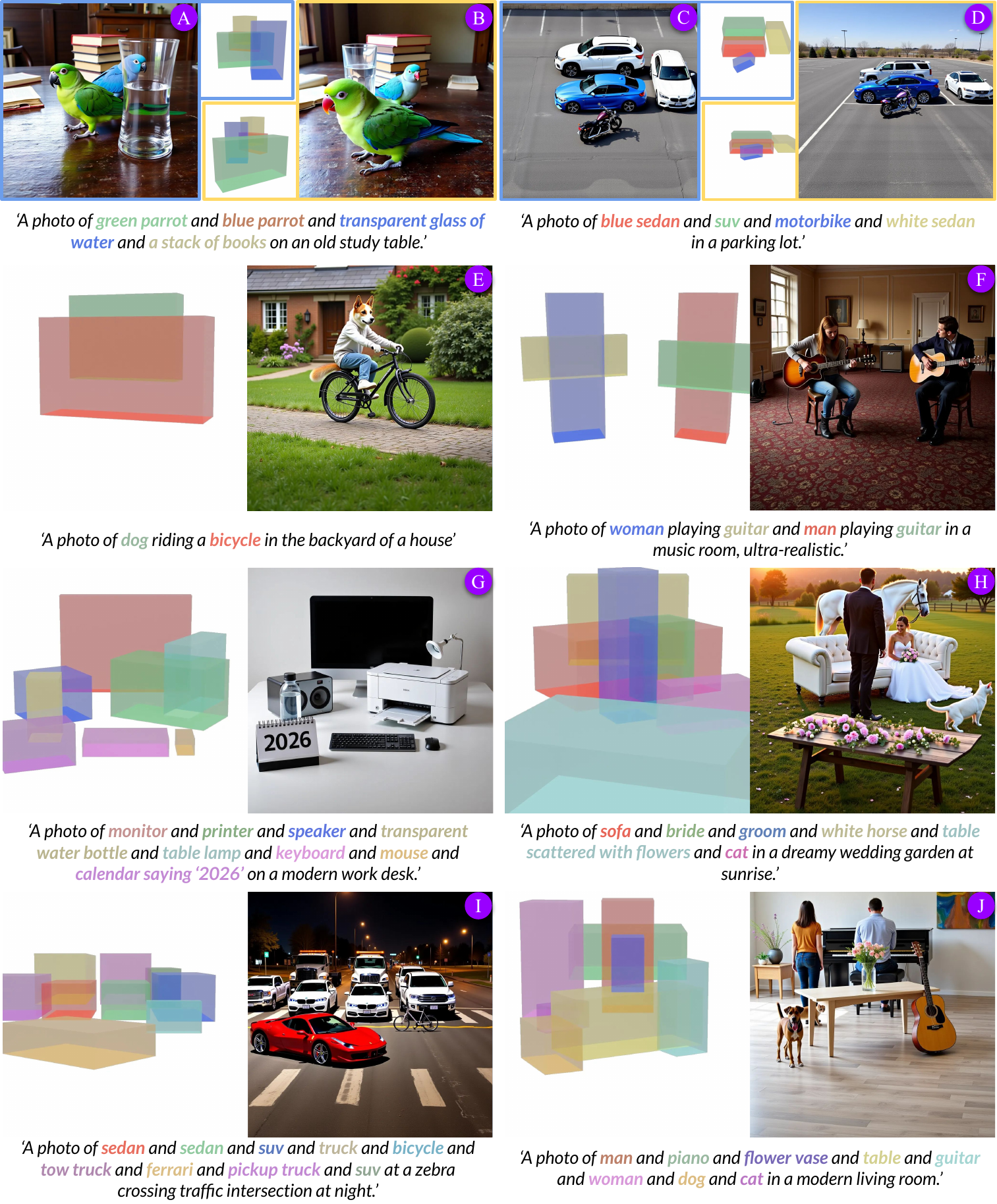}
   \caption{\textbf{Qualitative results:} Our method is able to precisely follow 3D scene layouts, with high occlusion consistency. Our approach preserves the prior of text-to-image model, as evident from capabilities like see-through transparent objects (A,B,G,J), text rendering (G) and inter-object interactions (E,F). Additionally, our method enables control over viewpoint of generated image (C,D). Despite being trained on layouts with only upto 4 objects, our method is able to generalize to complex scenes with many objects (G,H,I,J).} 
   \label{fig:qualitative_results}
\end{figure*}

%% file: sec/4_experiments.tex
\vspace{-2mm}

\section{Experiments}
\vspace{-1mm}

\begin{figure}[t]
    \centering
    \includegraphics[width=1.0\linewidth]{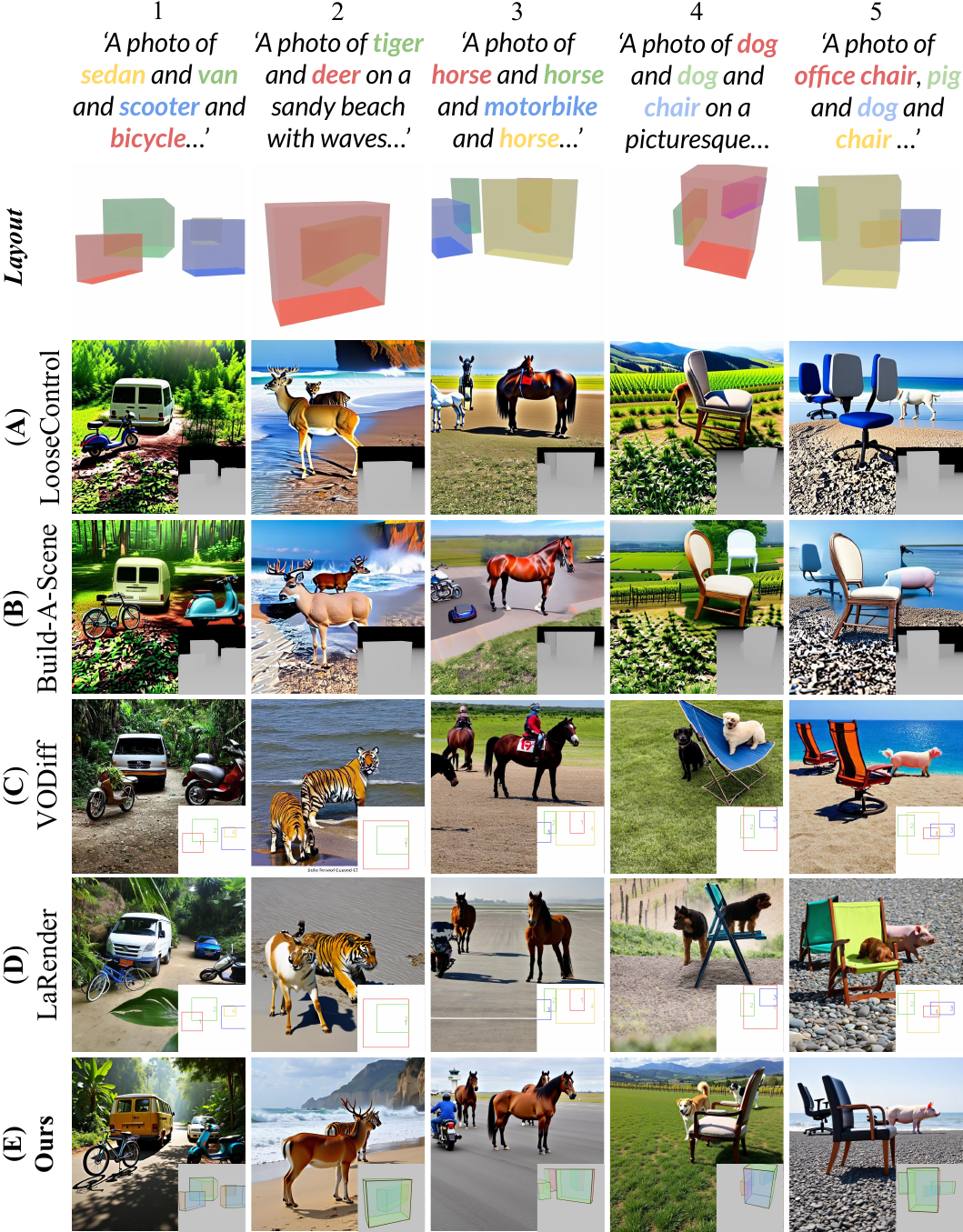}
    \caption{\textbf{Qualitative comparison:} We compare against works on \textbf{3D layout control}: \textit{LooseControl}~\cite{loosecontrol} and \textit{Build-A-Scene}~\cite{eldesokey2024build}, and on \textbf{occlusion control}: \textit{LaRender}~\cite{zhan2025larender} and \textit{VODiff}~\cite{liang2025vodiff}.}  
    \label{fig:qual_comparison}
    \vspace{-5mm}
\end{figure}

\subsection{Experimental setup}
\vspace{-1mm}
\textbf{Implementation details:} We use FLUX.1-dev~\cite{labs2025flux} as the text-to-image model. We train for $30K$ steps at a learning rate of $10^{-4}$, using a LoRA rank of $128$. A detailed implementation report can be found in appendix~\cref{sec:implementation_details}. 

\vspace{1mm}
\noindent
\textbf{Evaluation dataset:} Accurate evaluation of occlusion-aware 3D control requires a benchmark of paired images and 3D bounding box annotations that exhibit 1) diverse object configurations 2) challenging occlusion scenarios, and 3) wide range of camera viewpoints. To facilitate this, we introduce \textbf{3D} Control with \textbf{Oc}clusions benchmark, \textit{3DOc-Bench}, a dataset with $500$ samples of paired 3D bounding-box layouts, rendered images, and scene text prompts. We construct the benchmark in Blender~\cite{blender} by placing 3D assets on a ground plane and procedurally varying object arrangements and camera poses to produce strong occlusions while preserving a minimum visible area for each object. We will release the benchmark for future research in occlusion-aware generation. Detailed benchmark statistics are provided in the appendix~\cref{sec:appendix_dataset}.

\vspace{1mm}
\noindent \textbf{Evaluation metrics:} \label{subsec:metrics}
We measure the models' performance for layout adherence, text-to-image alignment, and image quality. For text-to-image alignment, we use CLIP image-text similarity, and for image quality, we use Kernel Inception Distance (KID)~\cite{bińkowski2018demystifying}. Evaluating 3D layout adherence using a single metric is challenging, as the generated scene may not conform to the metric depth specified by the 3D bounding-box layout. 
To this end, we compute three metrics that in unison effectively evaluate 3D layout adherence. Specifically, we compute 2D bounding box adherence, relative visibility order and 3D orientation consistency. (1) For evaluating 2D layout adherence, we first obtain object masks by combining 2D layouts with Segment Anything~\cite{kirillov2023segment}. Next, we compute CLIP similarity between the object masks and textual object descriptions, leading to CLIP \textit{objectness} score. We aggregate this objectness score to evaluate the 2D layout adherence (2) For evaluating relative visibility order, we adopt a similar method as~\cite{huang2025t2i}: we estimate per-pixel depth~\cite{yang2024depth} and obtain object depth estimates by averaging the depth within each object mask. Since all objects may not be present in the generated output, we use previously defined objectness score to filter out object masks. Finally, we compare relative depth ordering of each object pair against the ground-truth ordering, assigning a score of 1 if the ordering is correct and 0 otherwise. We aggregate this score over all such pairs 3) For assessing orientation accuracy, we employ OrientAnything~\cite{wang2024orient} to estimate object orientations using filtered object segments, and compute mean absolute error against ground truth.

\noindent \textbf{Baselines:} We compare our method with state-of-the-art works in \textbf{3D layout control}: \textit{LooseControl}~\cite{loosecontrol} and \textit{Build-A-Scene}~\cite{eldesokey2024build}. LooseControl uses layout depth maps to condition a diffusion model for scene layout control, while Build-A-Scene is an inference time method that uses pretrained LooseControl checkpoint. For fair evaluation, we train LooseControl on our dataset, and use the checkpoint to evaluate both methods. We also consider works on~\textbf{orientation control}, \textit{Compass Control}~\cite{parihar2025compass} and \textit{ORIGEN}~\cite{min2025origen}, though they do not support 3D object placement, hence not directly relevant. We compare against them in appendix~\cref{sec:extra_baselines}. We further evaluate against \textbf{occlusion control} methods,  \textit{LaRender}~\cite{zhan2025larender} and \textit{VODiff}~\cite{liang2025vodiff}. These methods decompose an image into 2D object layers to manage visibility ordering.

\begin{table}[t]
\centering
\resizebox{\columnwidth}{!}{%
\begin{tabular}{@{}cccccc@{}}
\toprule
\textbf{Baselines} & depth ord.$\uparrow$ & obj. score$\uparrow$  & angular err.$\downarrow$ & text align.$\uparrow$ & KID$(\times10^{-3})\downarrow$\ \\ \midrule 
VODiff~\cite{liang2025vodiff} & $0.68$ & {19.70}  & $92.73$  & $29.51$ & $15.40$       \\
LooseControl~\cite{loosecontrol}  & $0.82$ & $20.02$  & $89.88$  & $28.43$ & $14.32$      \\
Build-A-Scene~\cite{eldesokey2024build} & $0.89$ & $21.0$  & $91.62$  & $28.05$ & $20.12$        \\
LaRender~\cite{zhan2025larender} & $1.02$ & $21.83$ & $89.63$ & $30.20$ & $13.46$       \\
\textbf{Ours} & $\mathbf{1.46}$ & $\mathbf{22.86}$  & $\mathbf{47.92}$  & $\mathbf{31.87}$ & $\mathbf{5.43}$   \\ 
\bottomrule
\end{tabular}%
}
\vspace{-1mm}
\caption{\textbf{Quantitative comparison:} We compute (a) depth ordering, which reflects 3D location and occlusion consistency, (b) CLIP \textit{objectness} score, which indicates layout adherence and object fidelity (c) angular error, which indicates orientation correctness (d) image-text prompt alignment using CLIP~\cite{radford2021learningtransferablevisualmodels}, and (e) KID~\cite{binkowski2018demystifying}, which measures image fidelity.} 
\label{tab:quant_compare}
\vspace{-4.0mm} 
\end{table}

\subsection{Results}
\noindent \textbf{Qualitative:} We present our qualitative results in~\cref{fig:qualitative_results}. Our method is able to generate realistic scenes with intricate inter-object overlaps. It effectively preserves the prior of the base text-to-image model, evident from capabilities like see-through transparent objects (A,B,G,J) and text rendering (G). Additionally, our method enables control over viewpoint of generated image (C,D). Despite being trained on layouts with only upto 4 objects, our method is able to generalize to complex scenes with many objects (G,H,I,J). Even though our synthetic data consists of rigid objects in fixed canonical poses, our method is able to generate diverse poses such as sitting (H,J) and cycling (E). The model generates natural inter-object interactions (dog \textit{riding} bicycle in E, person \textit{playing} guitar in F), even though our synthetic data does not contain such interactions.  Further, it generalizes strongly to out-of-domain objects. Notably, our training dataset does not contain any musical instruments (F,J), electronic devices (G), transparent object (A,B,G,J) or books (A,B), but our model is able to effectively generalize to them. 

\vspace{1mm}
\noindent \textbf{Baseline comparisons:}\label{baseline_comparisons} We present results in~\cref{tab:quant_compare,fig:qual_comparison}. \textbf{3D scene control}: LooseControl~\cite{loosecontrol} fails to handle complex occlusions, as layout depth fails to represent occluded objects (see~\cref{fig:qual_comparison} A1,3-5). Additionally, the objects are generated in incorrect locations, due to lack of binding (A1,3), also reflected in low objectness-score (see~\cref{tab:quant_compare}). Build-A-Scene~\cite{eldesokey2024build} uses multiple generation and inversion cycles to sequentially add objects to the scene. While this improves upon layout adherence and occlusion consistency compared to LooseControl~\cite{loosecontrol}, it leads to inversion artifacts (B2-3,5), and hence worse KID value. The sequential generation also leads to lack of coherence in the generated scene (B4), since initial generations are independent of final scene layout. Both the methods fail to provide precise orientation control, since layout depth maps can only encode orientation upto $180^\circ$ flip, leading to high angular error. In contrast, our method is able to generate coherent images with precise 3D layout and orientation control. \textbf{Occlusion control:} \textit{LaRender}~\cite{zhan2025larender} and \textit{VODiff}~\cite{liang2025vodiff} rely on 2D layouts as conditioning input, which fail to discertain exact object arrangements. For instance, in~\cref{fig:qual_comparison} (C4, D4-5), the object is generated on `\textit{top} of the chair', against the intended configuration `\textit{behind} the chair'. In contrast, our OSCR representation is 3D aware, hence offers more precise control than 2D layouts. In case of large overlap between 2D bounding boxes in layout, baseline methods often fail to generate occluded objects (C1,3-4, D3-4) in contrast to SeeThrough3D, which can generate very occluded objects accurately (E).

\begin{figure}[h!]
    \centering
    \includegraphics[width=0.9\linewidth]{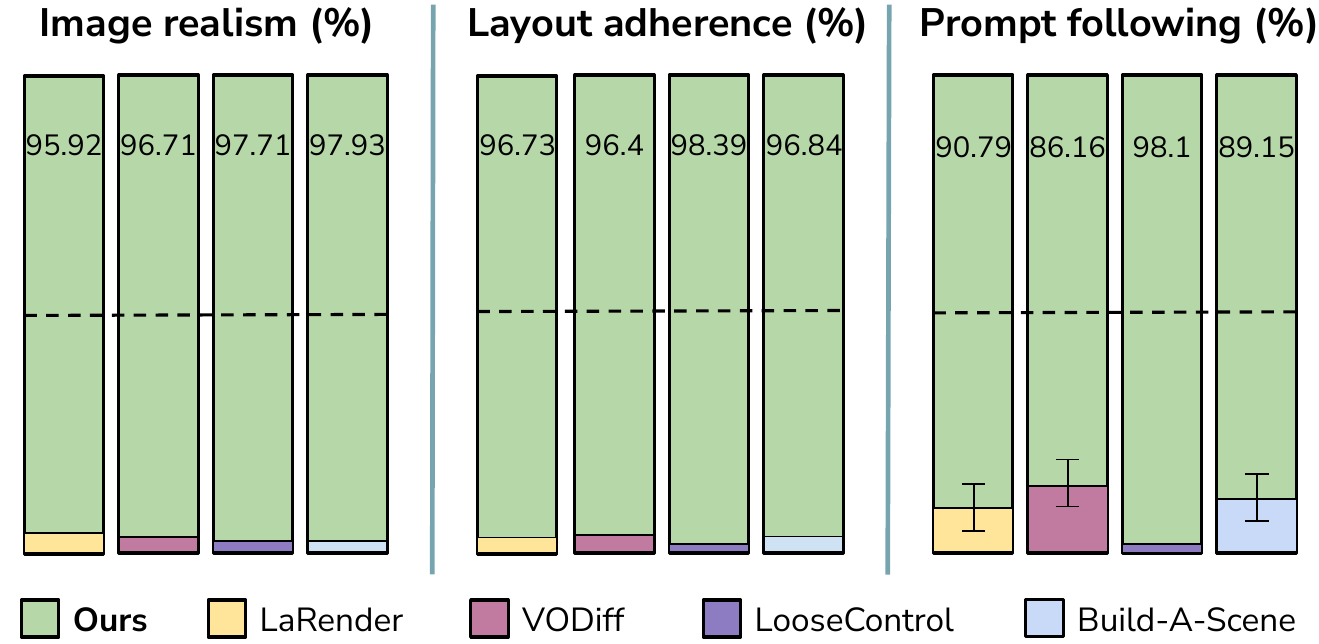}
    \caption{\textbf{User study:} Each bar indicates the $\%$ of times our method's output was preferred over the baseline, for each category.}
    \label{fig:user_study}
    \vspace{-5mm} 
\end{figure}

\noindent \textbf{User study:} We conducted an A/B user study where 60 participants were asked to choose between output of our method and a randomly chosen baseline. We evaluate a) image realism, b) layout adherence, and c) text prompt alignment. Results highlight high preference for our method in all evaluation categories (see~\cref{fig:user_study}). 

\subsection{Personalization} 
\begin{figure}[h!]
  \centering
   \includegraphics[width=1.0\linewidth]{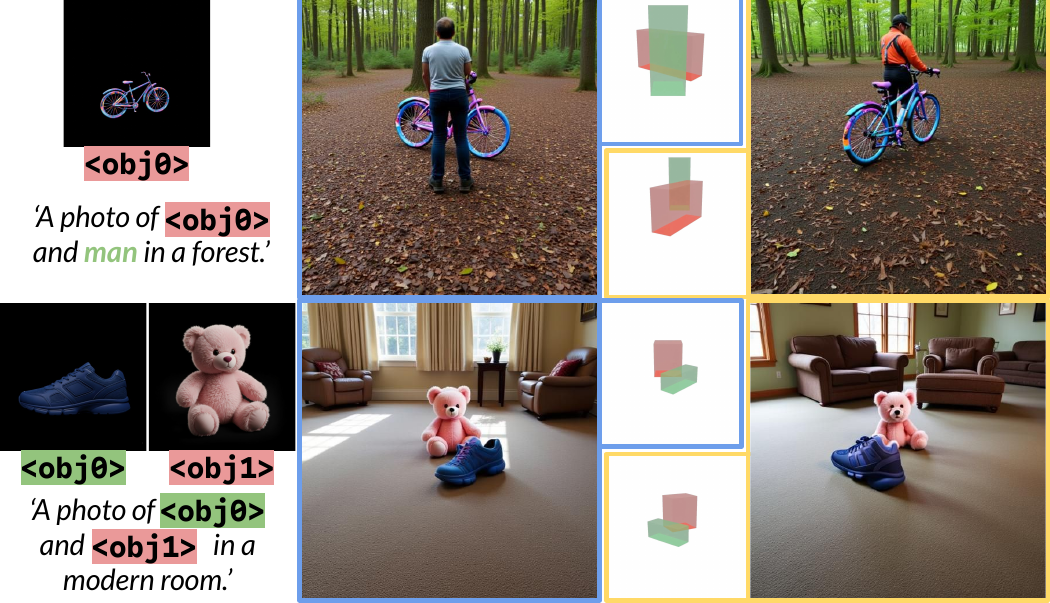}
   \vspace{-6mm}
   \caption{\textbf{Personalization:} Our method can be extended for personalized 3D control using reference image of an object.}
   \label{fig:personalization}
   \vspace{-1mm}
\end{figure}

\vspace{-1.5mm}
We show personalization results in~\cref{fig:personalization}. We adapt our training dataset for personalization by applying textures to 3D assets, and using this textured object as reference image. Further details and results are in appendix~\cref{sec:personalization_supplement}. 

\subsection{Ablations} 
\begin{figure}[!t] 
    \centering
    \includegraphics[width=1.0\linewidth]{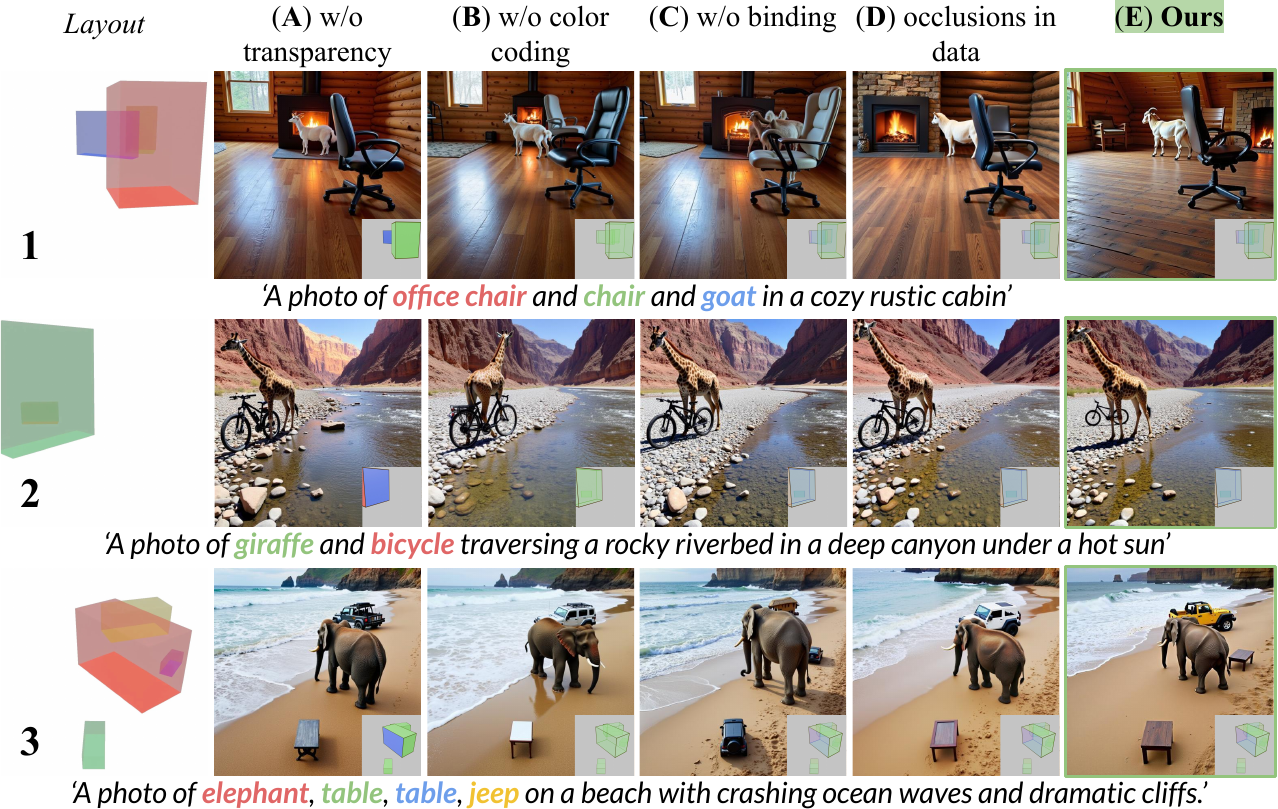}
    \vspace{-5mm}
    \caption{\textbf{Ablations:} We ablate upon key aspects of OSCR representation, our binding mechanism and data preparation strategy.} 
    \label{fig:ablations}
\end{figure}

We study the impact of key design choices, with results shown in \cref{fig:ablations,tab:ablations}. Box transparency plays a crucial role in the effectiveness of the OSCR representation, enabling reasoning about occluded objects and relative depth. Color-coding the box faces helps encode orientation and significantly reduces angular error (see \cref{tab:ablations}). Interestingly, opaque boxes yield the best orientation accuracy due to a clearer color signal. The attention-based binding is essential for layout adherence—without it, objects appear at incorrect locations (see \cref{fig:ablations}, 1C and 3C), resulting in lower objectness score. Finally, filtering out overly simplistic layouts in data improves performance. 

\vspace{-2mm}
\label{subsec:ablations}

\begin{table}[!t] 
\centering
\resizebox{\columnwidth}{!}{%
\begin{tabular}{@{}cccccc@{}}
\toprule
\textbf{Ablations} & depth ord.$\uparrow$ & obj. score$\uparrow$  & angular err.$\downarrow$ & text align.$\uparrow$ & KID$(\times10^{-3})\downarrow$\ \\ \midrule 
w/o transparency & $1.20$ & $21.67$    & $\mathbf{46.15}$ & $31.39$ & $5.90$ \\
w/o color-coding & $1.36$ & $22.23$ & $88.77$ & $31.57$ & $5.93$ \\
w/o binding & $0.98$ & $20.45$ & $57.44$ & $31.61$ & $6.35$      \\
w/o hard data & $1.24$ & $21.89$ & $49.73$ & $31.32$ & $6.34$ \\ 
\textbf{Ours} & $\mathbf{1.46}$ & $\mathbf{22.86}$  & ${47.92}$  & $\mathbf{31.87}$ & $\mathbf{5.43}$   \\ 

\bottomrule
\end{tabular}%
}
\vspace{-2mm}
\caption{\textbf{Quantitative results of ablative experiments.}} 
\label{tab:ablations}
\vspace{-5mm} 
\end{table}

%% file: sec/5_conclusion.tex
\section{Conclusion}
\vspace{-1.0mm}
We present SeeThrough3D, a model for occlusion aware 3D layout control. We introduce OSCR, an occlusion aware 3D scene representation. We show that our approach can faithfully model heavy occlusion scenarios, while preserving strong text-to-image prior of the model. Despite training on limited synthetic data, it exhibits strong generalization capabilities. We perform evaluations to show that our method outperforms existing baselines, and also ablate upon key design choices, providing useful insights for future research. While effective in layout adherence, our method does not preserve image consistency under layout changes. A future direction is to address this by using editing.

\section{Acknowledgements}
\vspace{-1.0mm}
We thank Harshavardhan P., Ayan Kashyap, Vansh Garg, Jainit Bafna, Abhinav Raundhal, Varun Gupta, Shivank Saxena, Akshat Sanghvi and Aishwarya Agarwal for helpful discussions and reviewing the manuscript. 

%% file: sec/X_suppl.tex


\appendix 

\clearpage
\setcounter{page}{1}

\section{Overview}
\label{sec:overview}

This appendix provides additional analysis, details about the dataset and model implementation, experimental discussion referenced in the main paper and extended qualitative results and comparisons. To skim over this material, the reader is advised to go through the figures and captions, which have been endowed with sufficient detail to understand the key content.


\section{Dataset}
\label{sec:appendix_dataset}

\VA{
Aims of this section
\begin{itemize}
\item a visualization of dataset preparation and filtering, show how we filter our bad examples in our Blender rendering process. 
\item a discussion of the augmentation pipeline, CLIP filtering. 
\item dataset statistics: camera elevation, object locations, occlusions.  
\item reference to the list of prompts.
\item a figure which has examples of data. 
\item discussion of personalization dataset preparation, with examples.  
\end{itemize}
}

\subsection{The rendering pipeline} 
\label{subsec:rendering_pipeline}
We collect 39 assets from Objaverse~\cite{objaverse} and SketchFab~\cite{spiess2024sketchfab} repositories from the internet. However, these assets are not aligned, making it difficult to define a canonical orientation for the objects. Hence, we align these assets manually in Blender~\cite{blender} to ensure that their canonical front directions are aligned with the +Y axis. We further scale each asset to match relative real-world dimensions, for example, the size of jeep is smaller than an elephant, the scale values were obtained using Gemini $2.5$ Pro~\cite{comanici2025gemini}. We place the aligned assets in a Blender environment (upto 4 objects per scene) and add a virtual camera to render the scene from a given viewpoint. Specifically, we define a hemispherical region around the origin of a fixed radius $R$, within which all objects are placed. The camera lies at the surface of the hemisphere, always pointing towards the origin.

However, randomly placing the assets and the camera might result in unnatural-looking compositions, such as those where objects are colliding with each other. Additionally, as described in the ablations section (main paper), a key requirement is that the objects must be heavily occluded to ensure optimal training of our model. To cater to the above requirements, we adopt a procedural generation, where the scene configuration (camera and object placements) is first randomly sampled from a uniform distribution of parameters with some predefined constraints. This is followed by a filtering logic to remove the poor-quality examples.

\noindent \textbf{Filtering based on occlusion:} We filter the rendered scenes according to the extent of occlusion, to ensure heavy occlusion scenarios. For this, we require a metric to measure the extent to which an object is occluded. Therefore, we define a visibility ratio $x$, which is the ratio of visible area $\mathbf{v}$ of the object to the total area $\mathbf{a}$ of the object. The values $\mathbf{v}$ and $\mathbf{a}$ are measured using object segmentation masks, obtained through Blender~\cite{blender}. We filter out cases where $x > 0.7$ for all objects in the scene, i.e., no object is occluded \textit{enough}. Similarly, we filter out cases where $x < 0.3$ for any object, to ensure that each object is adequately visible in the image.  

\noindent \textbf{Filtering based on object size:} We filter out cases where an object is too small or too large, to avoid unnatural-looking images. We filter based on the largest side of 2D object bounding boxes in the renderings. Specifically, we ensure that the largest side of the 2D bounding box must be within $0.125$ and $0.750$ of the image size.

\begin{figure*}[h]
  \centering
   \includegraphics[width=1.0\linewidth]{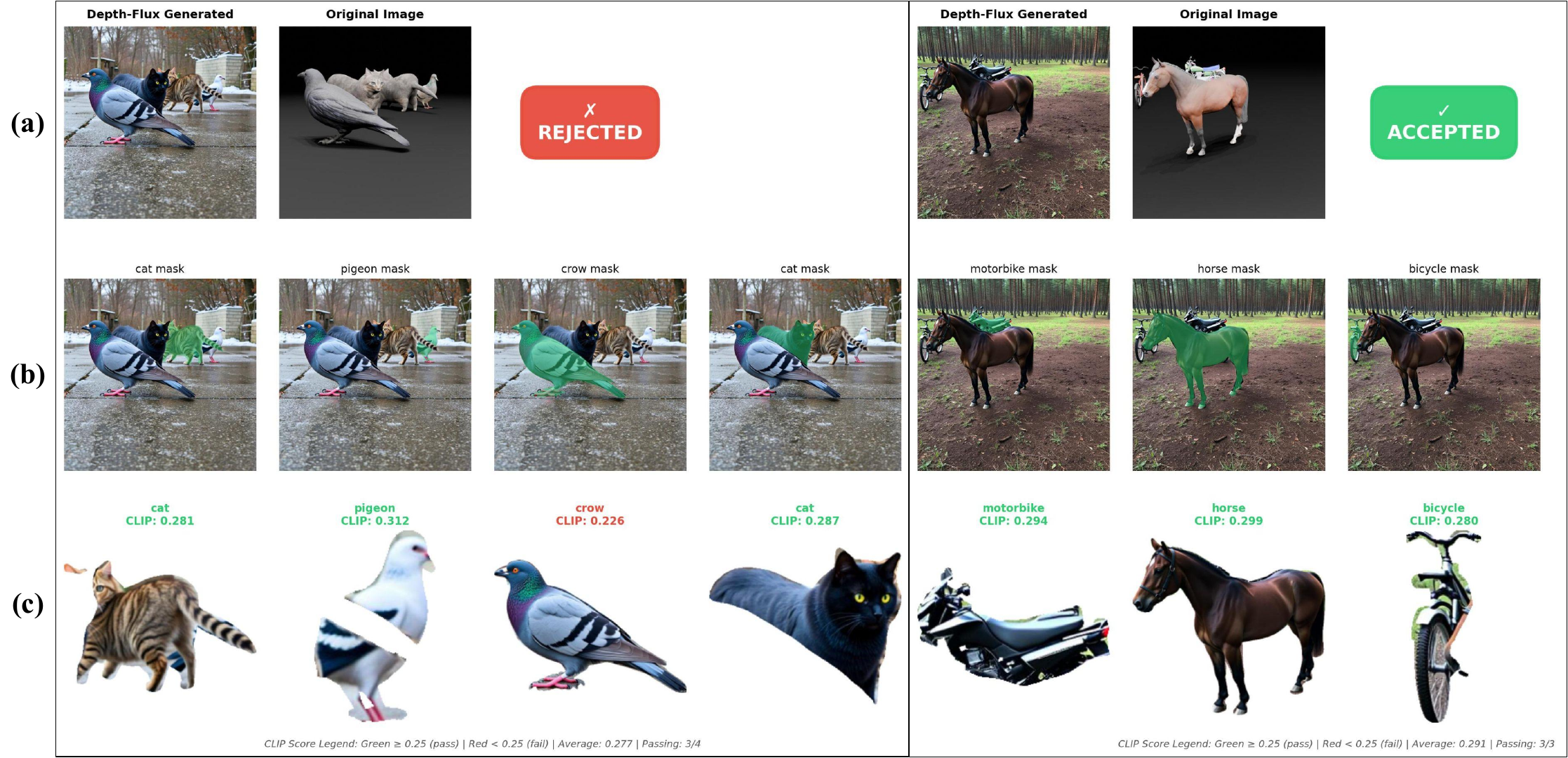}
   \caption{\textbf{CLIP filtering on augmentations:} We use a depth-to-image model (FLUX.1-Depth-dev~\cite{labs2025flux}) to generate realistic augmentations of the rendered images (a). However, the depth-to-image model occasionally misaligns objects with their intended depth regions, causing incorrect placements. For example, on the left pane, the depth-to-image model incorrectly generates a pigeon in place of a crow according to original layout. We mitigate such issues by applying object-level CLIP-based filtering~\cite{radford2021learningtransferablevisualmodels} to retain only those augmentations that adhere to the original layout. For this, we first obtain the object segmentation masks using Blender~\cite{blender}, and use these to obtain cropped object segments in the augmented image (b). Next, we compute CLIP similarity between these object segments and corresponding text description (\eg{} cat, pigeon, etc.), as shown in (c). If any object has a CLIP score less than the threshold value of $0.25$, the augmentation is filtered out. High CLIP scores for all object segments (as shown on the right pane) indicates accurate layout adherence, and these images are included in the training dataset.}
   \label{fig:augmentation_pipeline} 
   \vspace{5mm}
\end{figure*}

\subsection{Augmentations}
Training solely on these rendered images from Blender risks overfitting to synthetic backgrounds~\cite{parihar2025compass,cont-words}, due to limited realism and lack of diversity in object appearance and backgrounds. Since creating highly varied 3D scenes is an expensive process, we adopt a scalable alternative. We generate realistic augmentations for the rendered images that follow the same layout but are rich in terms of appearance diversity. For each rendered image, we extract its depth and pass it through a depth-to-image generation pipeline (FLUX.1-Depth-dev)~\cite{labs2025flux} to synthesize realistic images that preserve the same spatial layout. 

\begin{figure*}[t]
  \centering
   \includegraphics[width=1.0\linewidth]{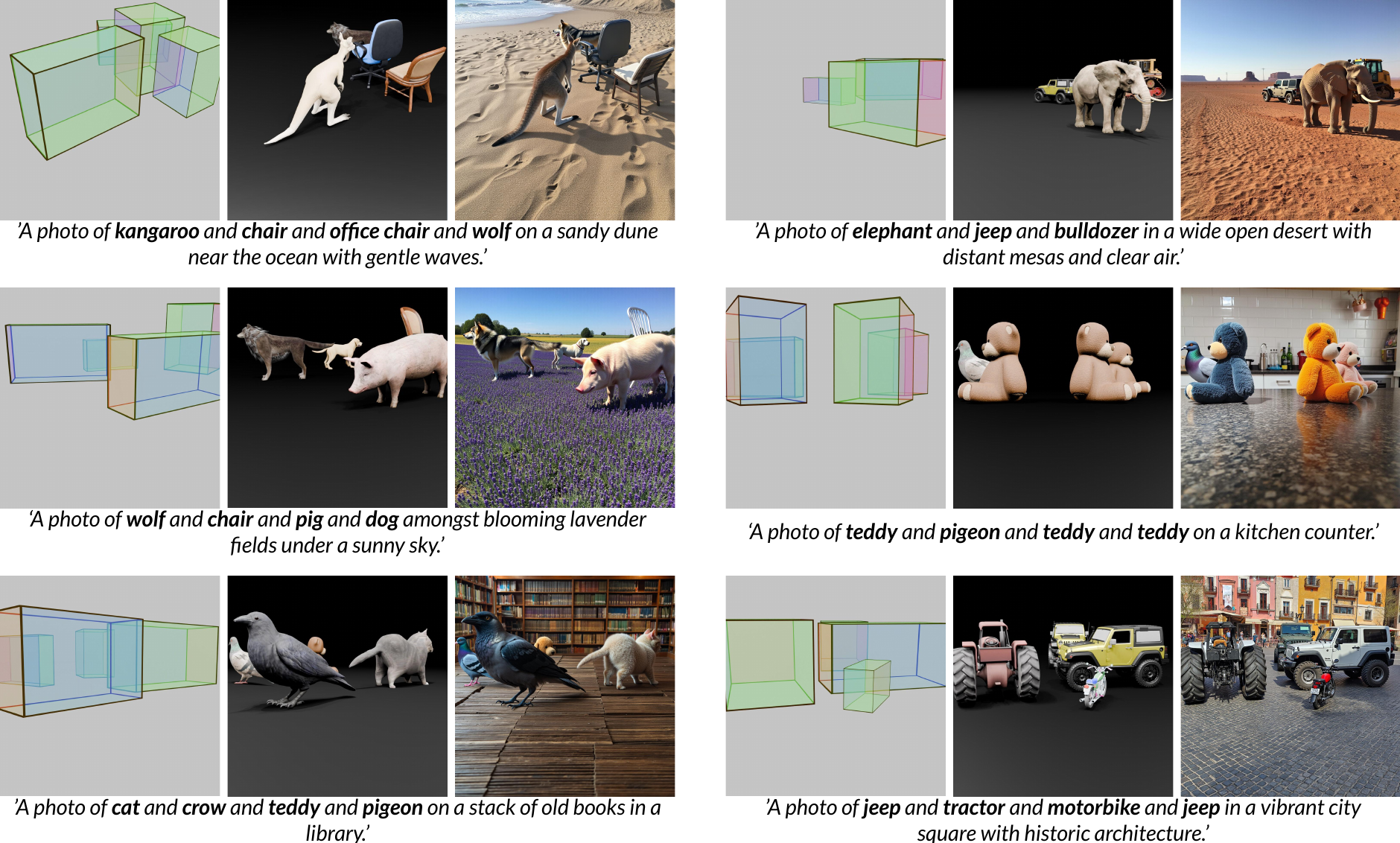}
   \caption{\textbf{Samples from our training dataset:} We create scenes in Blender~\cite{blender} by placing 3D assets in controlled configurations and defining the rendering camera viewpoint. The object arrangements and camera viewpoint are controlled to ensure strong inter-object occlusions, while ensuring that each object is sufficiently visible in the image. Along with the main image, we render the corresponding OSCR representation, which consists of color-coded translucent 3D bounding boxes of the objects. The rendered images are further augmented using a depth-to-image pipeline to obtain realistic images that follow the same layout as shown in Fig:\ref{fig:augmentation_pipeline}.}
   \label{fig:data_examples} 
\end{figure*}

\noindent 
\textbf{Filtering augmentation samples.}
Although this pipeline produces high-quality results, it occasionally misaligns objects with their intended depth regions, causing incorrect placements. For instance, on the left pane in~\cref{fig:augmentation_pipeline}(a), the depth-to-image model incorrectly generates a pigeon, instead of a crow, according to the original layout. We mitigate such issues by applying object-level CLIP-based filtering~\cite{radford2021learningtransferablevisualmodels} to retain only those augmentations that adhere to the original layout. For this, we first obtain the object segmentation masks using Blender~\cite{blender}, and use these to obtain cropped object segments in the augmented image (see~\cref{fig:augmentation_pipeline}(b)). Next, we compute CLIP similarity between these object segments and corresponding text description (\eg{} cat, pigeon, etc.), as shown in~\cref{fig:augmentation_pipeline}(c). If any object has a CLIP score less than the threshold value of $0.25$, the augmented image is filtered out. High CLIP scores for all object segments (as shown on the right pane in~\cref{fig:augmentation_pipeline}) indicate accurate layout adherence, and these images are included in the training dataset. We visualize some examples from our training dataset in~\cref{fig:data_examples}.

\subsection{Statistics}
\begin{figure*}[t]
  \centering
   \includegraphics[width=1.0\linewidth]{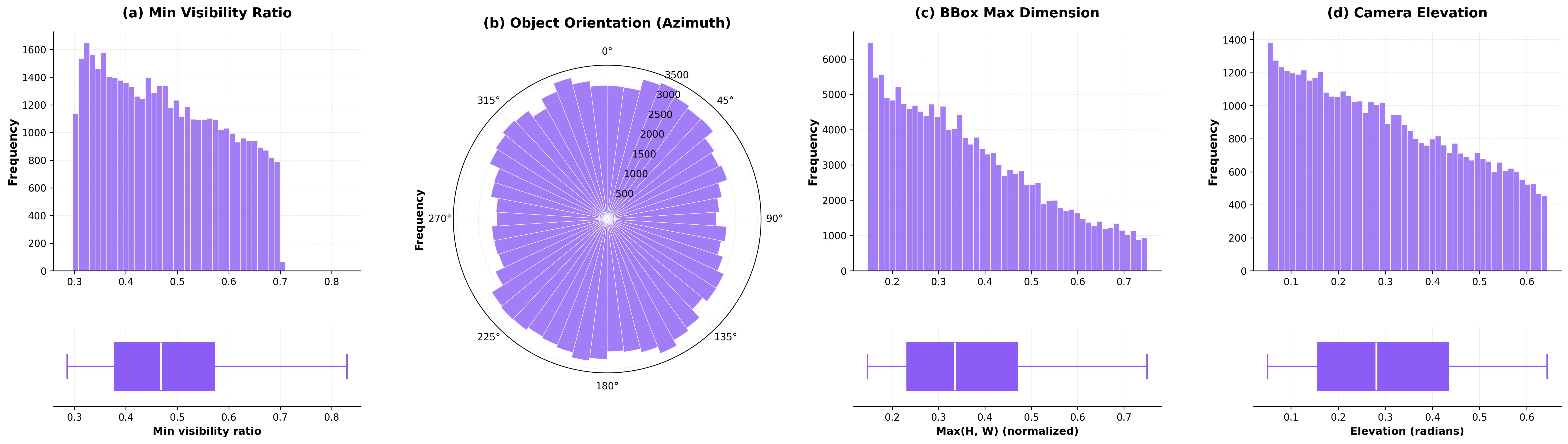}
   \caption{\textbf{Statistics of training dataset:} (a) We plot the distribution of minimum visibility ratio for any object in the scene. Since our filtering strategy favors heavy occlusion scenarios, we observe that there is a bias towards cases with low visibility ratios. (b) Next, we observe that the distribution of orientation values is roughly uniform, thus avoiding any unwanted biases. (c) Interestingly, we observe that the frequency of examples with large 2D bounding box dimension shows a decreasing trend. This is because smaller object sizes enable placement of multiple objects in a scene, while ensuring all of them are visible. (d) High camera shots tend to have weaker occlusions compared to low camera shots; for instance, there are very little inter-object occlusions in bird's-eye-view of a scene (high camera elevation). Since our data selection process favors high occlusion scenarios, renders with low camera are usually favored by the rendering pipeline algorithm, explaining the observed trend.} 
   \label{fig:training_stats} 
\end{figure*}
\begin{figure*}[t]
  \centering
   \includegraphics[width=1.0\linewidth]{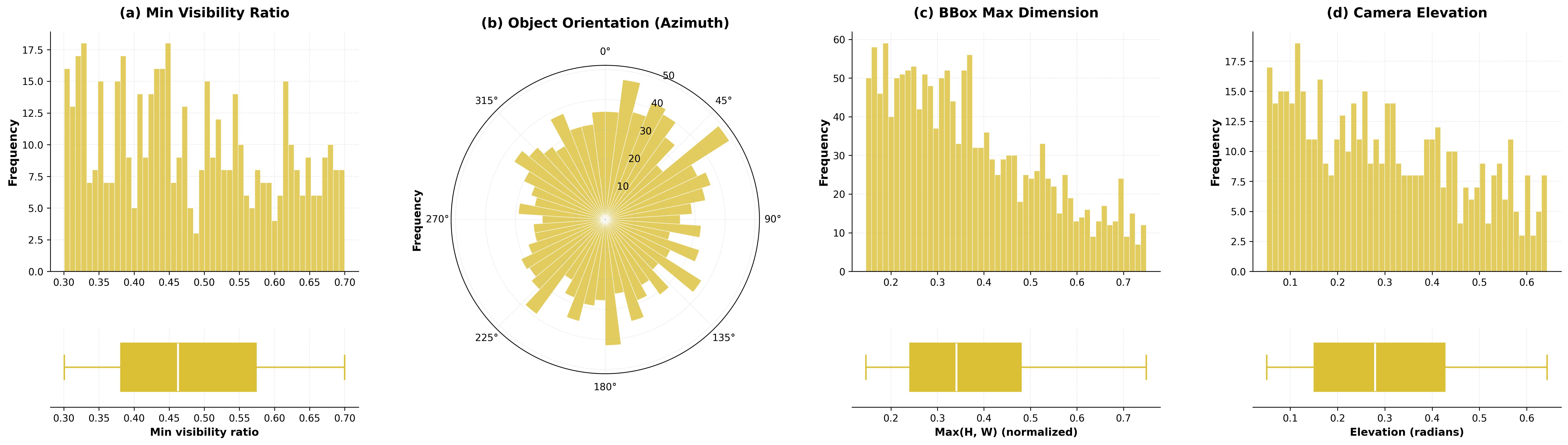}
   \caption{\textbf{Statistics of \textit{3DOcBench} evaluation benchmark:} Similar to the training dataset, we observe that the 3DOcBench evaluation benchmark contains (a) heavy occlusion scenarios, (b) roughly uniform distribution of  orientations, (c) higher frequency of smaller objects, measured using 2D bounding box dimension, and (d) large number of cases with low camera elevation and consequently high inter-object occlusion.} 
   \label{fig:eval_stats} 
\end{figure*}

We present various statistics of our training dataset in~\cref{fig:training_stats}. (a) We plot the distribution ofthe  minimum visibility ratio for any object in the scene. Since our filtering strategy favors heavy occlusion scenarios, we observe that there is a bias towards low visibility ratio cases. (b) Next, we observe that the distribution of orientation values is roughly uniform, thus avoiding any unwanted biases. (c) Interestingly, we observe that the frequency of examples with large 2D bounding dimensions shows a decreasing trend. This is because smaller object sizes enable the placement of multiple objects in a scene, while ensuring all of them are visible. (d) By common observation, high camera shots tend to have weaker occlusions compared to low camera shots. For instance, there are few occlusion scenarios in bird's-eye-view (high camera elevation) of a scene. Since our data selection process favors high occlusion scenarios, renders with low camera are favored by the rendering pipeline algorithm, explaining the observed decreasing trend.

\section{\textit{3DOcBench} benchmark details}
\VA{
Aims of this section
\begin{itemize}
\item the statistics of this benchmark. 
\end{itemize}
}

For constructing our evaluation benchmark, \textit{3DOcBench}, we use the same procedural generation to prepare the training dataset (see~\cref{subsec:rendering_pipeline}). Specifically, we construct scene layouts in Blender~\cite{blender} with the 3D assets and camera placed in random locations, and filter the layouts based on whether they meet the constraints of occlusion and object size (see~\cref{subsec:rendering_pipeline}). We present various statistics of the benchmark in~\cref{fig:eval_stats}. Similar to the training dataset, we observe that the 3DOcBench evaluation benchmark contains (a) heavy occlusion scenarios, (b) roughly uniform distribution of orientations, (c) higher frequency of smaller objects, measured using 2D bounding box dimension, and (d) a large number of cases with low camera elevation and consequently high inter-object occlusion.

\section{Overlapping regions}
\label{sec:attention_analysis}
\begin{figure*}[t]
  \centering
   \includegraphics[width=1.0\linewidth]{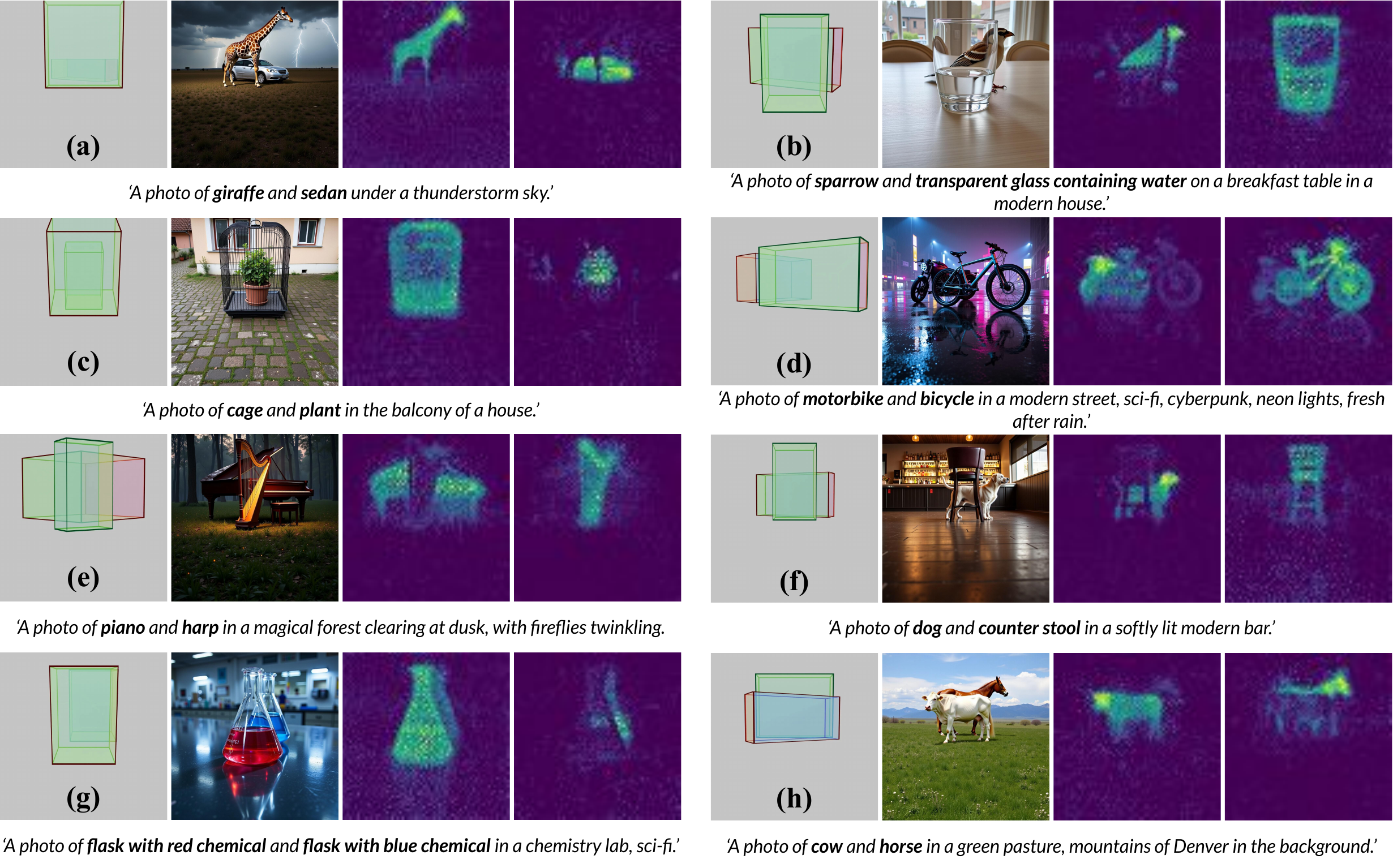}
   \caption{\textbf{Visualizing object disentanglement in latent space:} We use the layouts (shown in first frame) to condition our model, the outputs are shown in second frame. We store the intermediate attention maps from image tokens to object tokens in the text prompt, visualized in third and fourth images. Evidently, the attention maps reveal occlusion boundaries, and show some interesting patterns. Notably, in cases involving transparent objects like water (b) and chemical flask (g), the physically hidden regions of sparrow and flask respectively are visible in attention map. Even in case of semantically similar categories, such as cow and horse (a), flasks with differently colored chemicals (g), motorbike and bicycle (d) the attention is highly localized, with only minimal leakage.}  
   \label{fig:attn_viz_detailed} 
\end{figure*}

In the proposed object binding strategy, the OSCR tokens at the intersection of two rendered bounding boxes attend to all participating object tokens in the text prompt. However, at first glance, it seems that attending to multiple object semantics would lead to semantic bleeding and visual artifacts at object boundaries. Upon investigation, however, we found that the generated images feature sharp occlusion boundaries without object attribute mixing. To understand this, we visualize the attention maps between image and text tokens, and find that object features are segregated in the model's latent space. We analyse the attention maps through $8$ complex scene layouts (of two objects) with heavy occlusion scenarios in Fig.~\cref{fig:attn_viz_detailed}. As we can see, the attention maps clearly distinguish between the foreground and background objects with appropriate occlusions. This indicates the inherent model's capability in handling object occlusions, and our method provides a new interface to accurately generate such scenes, which is challenging to do with text alone.

\begin{figure*}[]
  \centering
   \includegraphics[width=1.0\linewidth]{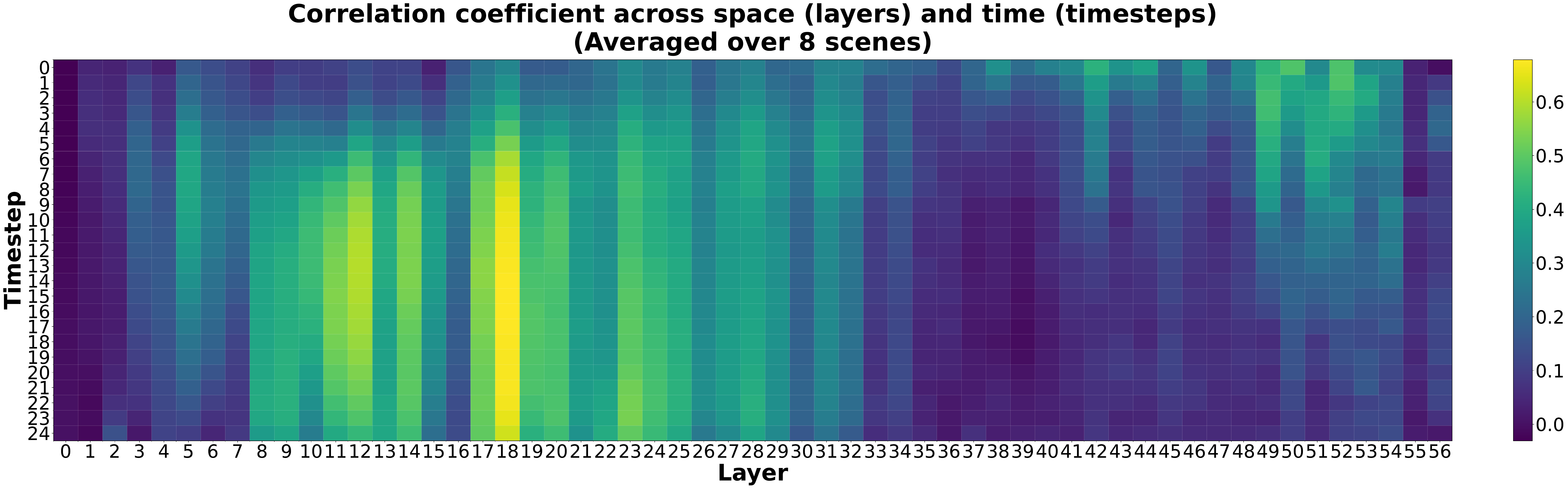}
   \caption{\textbf{Measuring spatial alignment of image to object attention using correlation coefficient (CC):} We create a dataset of 8 complex layouts containing strong occlusion scenarios (see~\cref{fig:attn_viz_detailed}). We obtain SeeThrough3D's outputs on these layouts, and store intermediate attention maps from image tokens to object tokens in the text prompt. Next, we run Segment Anything~\cite{kirillov2023segment} on the generated outputs to obtain object-level segmentation masks. Finally, we use correlation coefficient (CC) to measure alignment between the ground truth object segment and corresponding image to object attention maps. We compute the CC across space (layers) and time (denoising timesteps), and the obtained heatmap reveals interesting insights. For a given layer, timestep combination, a high CC value indicates strong spatial awareness. We observe that very specific layers in the DiT are spatially aware; early layers from 8 to 25, after which the spatial awareness decreases sharply. Secondly, the spatial properties in attention emerge after 5th denoising step (out of 25 steps) in these layers. The pattern of spatially aware layers is very irregular, indicating that different layers in the DiT contribute very differently to the generated image, consistent with findings from~\cite{avrahami2025stable}.}
   \label{fig:space_time}
\end{figure*}

\noindent
\textbf{Selecting layers for attention visualization.} We performed a simple analysis for choosing the appropriate layers for attention visualization. We use Segment Anything~\cite{kirillov2023segment} on the generated images followed by manual filtering to segment out individual object regions. Finally, we measure spatial alignment between image to object token attention and ground truth segmentation using correlation coefficient (CC). We analyze CC values across space (DiT layers) and time (denoising timesteps), results are visualized in~\cref{fig:space_time}. The results highlight that spatial alignment is high for very specific layers in the DiT, particularly for layers $11$ to $23$. Also, spatial alignment tends to emerge at around $5th$ denoising timestep (out of $25$ timesteps). We use the resulting CC matrix to filter top-$50$ (layer, timestep) combinations which show highest spatial alignment, and average image to text attention for each object. The results are visualized in~\cref{fig:attn_viz_detailed}.  

\section{Implementation details}
\label{sec:implementation_details}

\VA{
Aims of this section
\begin{itemize}
\item talk about the model details. 
\item talk about training details. 
\item talk about logistics, training time, inference time, etc. 
\end{itemize}
}
We build upon FLUX.1-dev~\cite{labs2025flux} as our base model. We patch it with $128$-rank LoRA adapters, applied on query, key and value projections in every attention layer. Additionally, we set the LoRA scale to $0$ for the text and image tokens to preserve the strong text-to-image prior of the base model~\cite{tan2024ominicontrol,tan2025ominicontrol2,zhang2025easycontrol}. We train our model with a learning rate of $10^{-4}$ using the AdamW optimizer for $30K$ steps with an effective batch size of $2$. The first $25K$ training steps use an image resolution of $512$, followed by a resolution of $1024$ for the next $5K$ steps. We found that such staged training helps improve realism in the generated images. The complete training takes around $9$ hours on $2X$ NVIDIA $H100$ GPUs (one image per GPU). Our implementation is based on PyTorch~\cite{paszke2019pytorch} and Hugging Face Diffusers~\cite{von-platen-etal-2022-diffusers} framework.

To enable personalization, we introduce an additional `subject' LoRA of the same rank (128) on the reference image tokens. Both the LoRA's are finetuned on our personalization dataset (see~\cref{sec:personalization_supplement}) for $7.5K$ iterations.

\section{Taking control with SeeThrough3D}
\begin{figure*}
  \centering
   \includegraphics[width=0.78\linewidth]{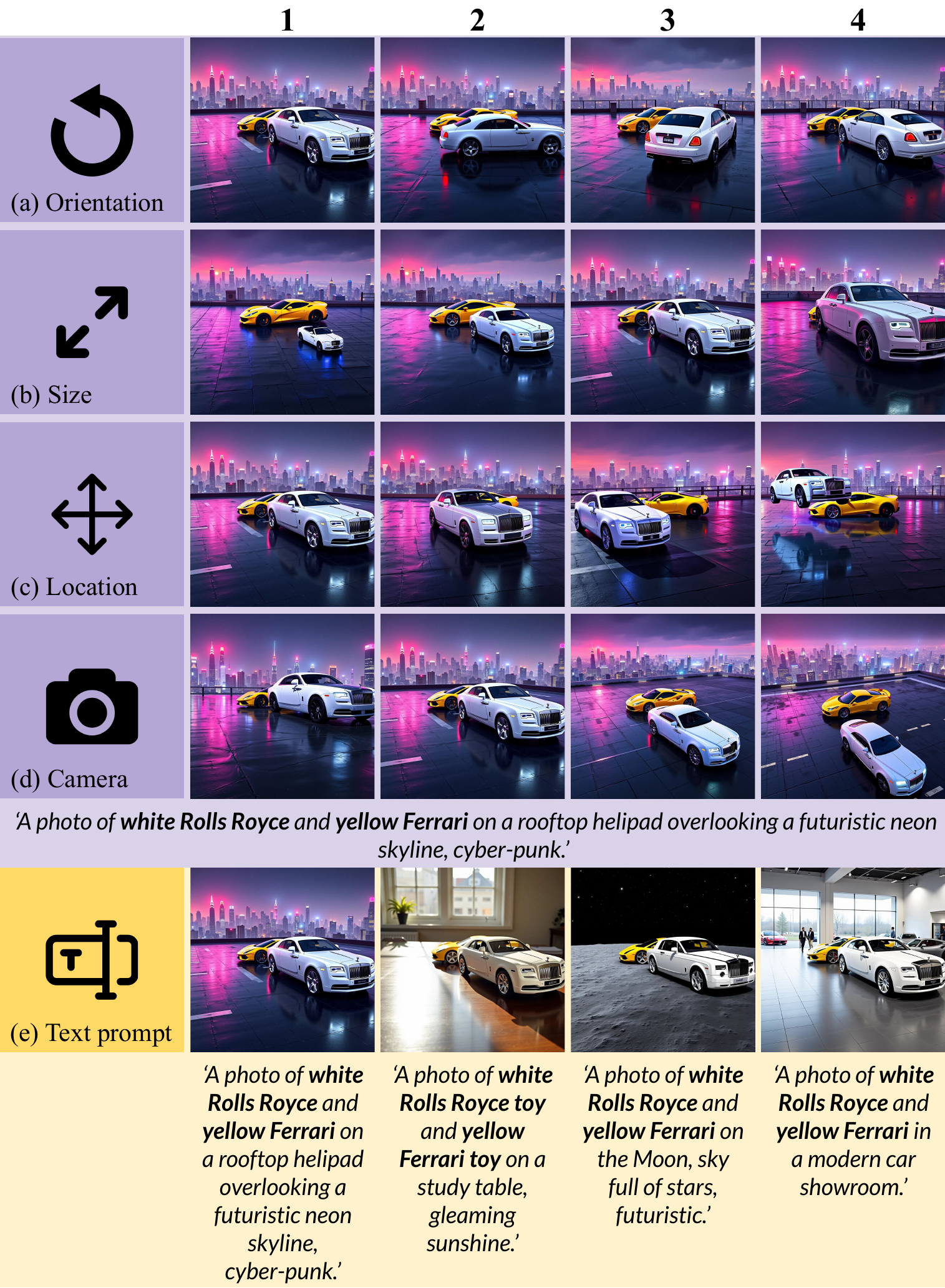}
   \caption{\textbf{Taking control with SeeThrough3D:} We demonstrate the individual controls that our approach offers over the scene composition, which includes 3D attributes such as (a) object orientation, (b) object size, (c) object location, (d) scene camera elevation, as well as (e) text prompt and object semantics, all while ensuring occlusion consistency. Notably, all the images in this figure were generated \textbf{using the same random seed}, highlighting the effectiveness of control. \textbf{Disentangled control:} In (a), (b) and (c), we are able to control the 3D attributes of one object (Rolls Royce), without altering the other object, indicating disentangled control. Notice how occlusion consistency is preserved even in case of heavy overlap (b4), when the white car has become very big, and in (d1), where the camera elevation is very low. The model is able to model interesting controls such as levitating objects (c4). Despite heavy overlaps, the \textbf{object attributes (`white Rolls Royce', `yellow Ferrari') remain correctly bound} to respective objects without attribute mixing, highlighting effectiveness of our binding mechanism.}  
   \label{fig:various_controls_a} 
\end{figure*}

\begin{figure*}
  \centering
   \includegraphics[width=0.78\linewidth]{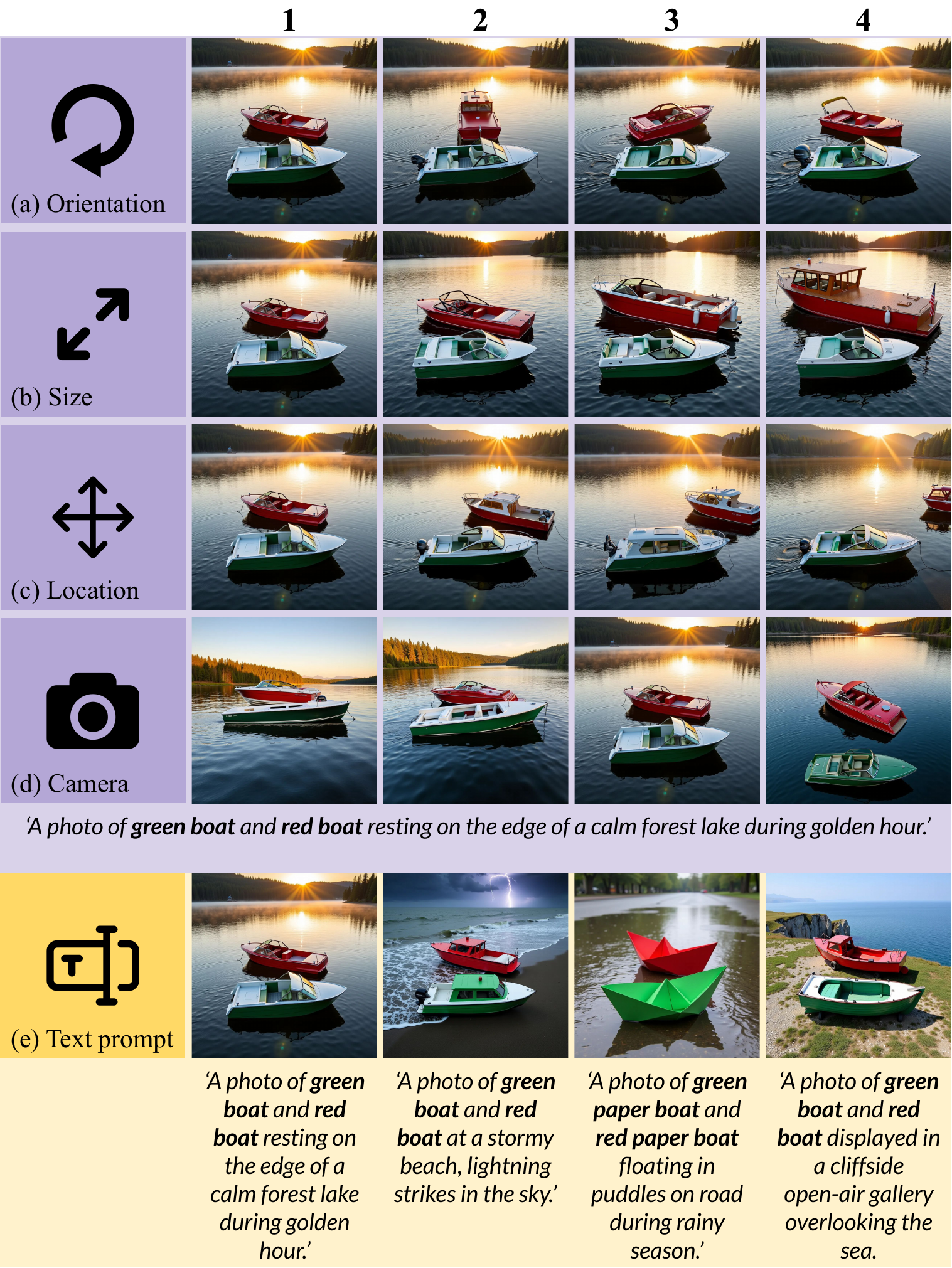}
   \caption{\textbf{Taking control with SeeThrough3D:} We demonstrate the individual controls that our approach offers over the scene composition, which includes 3D attributes such as (a) object orientation, (b) object size, (c) object location, (d) scene camera elevation, as well as (e) text prompt and object semantics, all while ensuring occlusion consistency. Notably, all images in this figure were generated \textbf{using the same random seed}, highlighting the effectiveness of control. \textbf{Disentangled control:} In (a), (b) and (c), we are able to control the 3D attributes of one object (red boat), without altering the other object, indicating disentangled control. Notice how occlusion consistency is preserved in challenging cases like (d1), where the camera elevation is very low. Despite heavy overlaps, the \textbf{object attributes (`green boat`, `red boat`) remain correctly bound} to respective objects without attribute mixing, highlighting effectiveness of our binding mechanism.} 
   \label{fig:various_controls_b} 
   \vspace{5mm}
\end{figure*}

OSCR representation encodes various 3D attributes of a scene, such as object orientation, size and location, along with camera viewpoint as well as occluded object regions. This enables SeeThrough3D to control all the properties in jointly. Additionally, since our method preserves the strong prior of the base text-to-image model, it can generate diverse visual appearance of both objects and background, solely through text prompt control. These diverse forms of control offered by SeeThrough3D are summarized in~\cref{fig:various_controls_a,fig:various_controls_b}. Note that \textbf{all the images in these figures are generated using the same random seed}, highlighting the effectiveness of control. Notably, the model is able to preserve occlusion consistency even despite heavy overlaps, such as low camera elevation (d1,~\cref{fig:various_controls_a,fig:various_controls_b}), (b4,~\cref{fig:various_controls_a}). These results indicate the preciseness of control enabled by our method, enabling various applications in design and architecture.  

\section{Additional baseline comparisons}
 \label{sec:extra_baselines}
\VA{
Aims of this section
\begin{itemize}
\item compare with baselines we have skipped in the main paper 
\end{itemize}
}

In the main paper, we have compared against \textbf{3D scene control} methods, \textit{LooseControl}~\cite{loosecontrol} and \textit{Build-A-Scene}~\cite{eldesokey2024build}. These methods are directly relevant to ours, since they enable control over 3D scene layout, including object placement, orientation and camera viewpoint. Here we compare against baselines which specifically allow control over 3D object orientation only, without controlling 3D object placement or camera viewpoint. We compare against two baselines, \textit{ORIGEN}~\cite{min2025origen} and \textit{Compass Control}~\cite{parihar2025compass}. ORIGEN enables control over object orientation using a one step generative model. Specifically, they perform initial noise optimization according to a reward function which penalizes the mismatch between orientation of the generated object and the input orientation angle. The generated object orientation that is measured using Orient Anything~\cite{wang2024orient}. However, they do not provide control over locations of the objects. Compass Control, on the other hand, enables control over object orientation along with 2D object layouts. They learn an adapter which maps object orientation to a per object \textit{compass} embedding. These embeddings are then used to condition the generative process through cross attention. The cross attention maps of compass and object tokens in prompt are constrained to respective 2D bounding boxes to enable disentangled orientation control for multi-object scenes. 

\begin{figure}[h]
    \centering
    \includegraphics[width=1.0\linewidth]{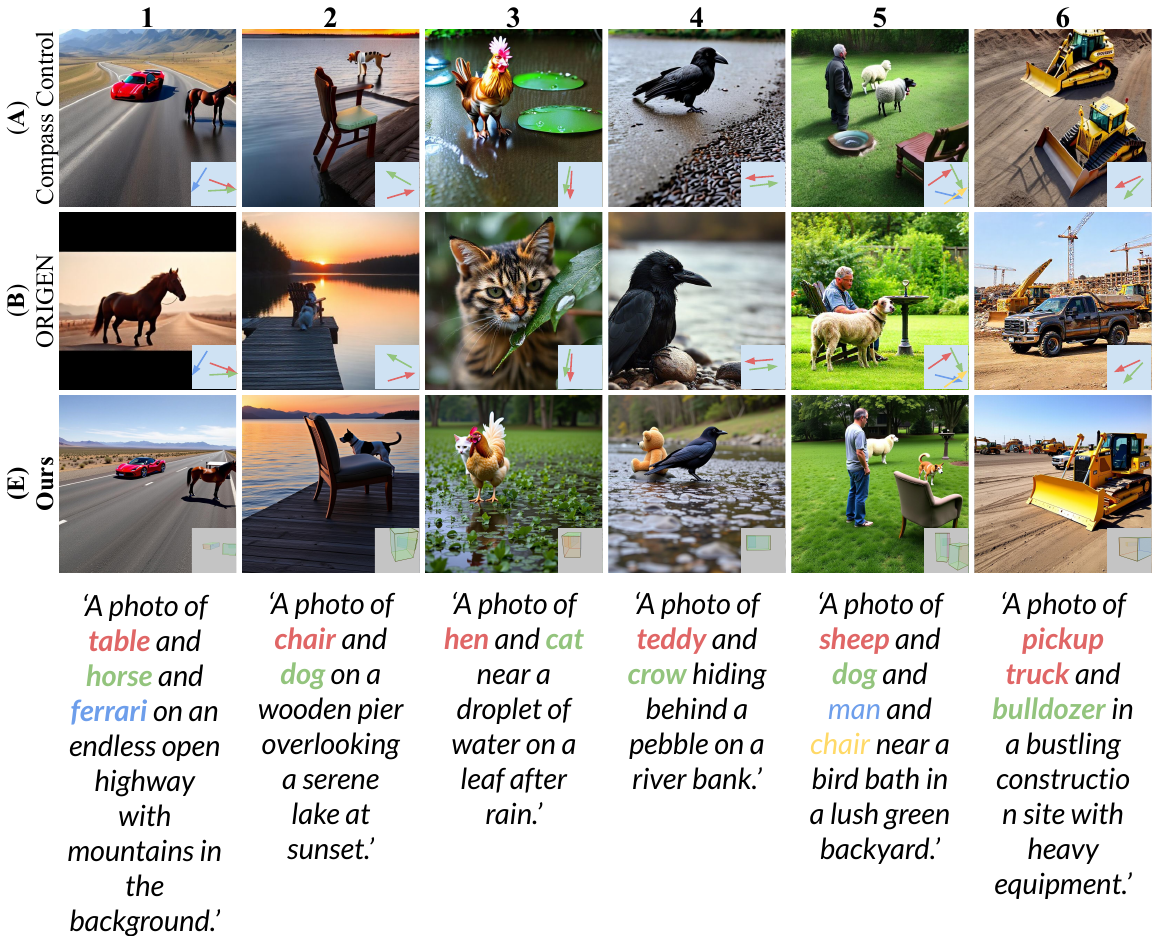}
    \caption{\textbf{Qualitative comparisons with additional baselines:} We present comparisons with baselines \textit{Compass Control}~\cite{parihar2025compass} and \textit{ORIGEN}~\cite{min2025origen}. We observe that Compass Control is not able able to handle complex occlusions (A1-4), and mixes object attributes in case of heavy overlaps (A5-6). ORIGEN fails to generate some objects in the scene (B1-6), and its outputs contain visual artifacts arising from poor noise optimization (B2). Additionally, ORIGEN is limited to one-step generative models, and hence suffers from low image fidelity. In contrast, our method is able to model complex occlusions (E1-6) without attribute mixing, indicating its effectiveness.}   
    \label{fig:qual_comparison_origcomp}
\end{figure}

\noindent \textbf{Analysis:} We present comparison results against these baselines in~\cref{fig:qual_comparison_origcomp,tab:quant_compare_supply}. Since ORIGEN~\cite{min2025origen} does not allow 2D layout control, it is not compatible with our quantitative evaluation that focuses on layout adherence (see \textbf{Evaluation metrics} in the main paper), and we only present qualitative comparisons. We observe that Compass Control is not able able to handle complex occlusions ($A1-4$), and mixes object attributes in case of heavy overlaps ($A5-6$), resulting in low objectness score. ORIGEN fails to generate some objects in the scene (B1-6). Additionally, its outputs contain artifacts arising from poor noise optimization (B2). Additionally, ORIGEN is limited to one-step generative models, and hence suffers from low image fidelity. In contrast, our method is able to model complex occlusions (E1-6) without attribute mixing, indicating its effectiveness.

\section{More on angular error evaluation} 
Two of our baselines, \textit{LooseControl}~\cite{loosecontrol} and \textit{Build-A-Scene}~\cite{eldesokey2024build} use layout depth maps as condition for text-to-image model. While providing 3D placement cues, the layout depth representation fails to capture precise 3D orientation, leading to poor orientation accuracies, as indicated by high angular error values in~\cref{tab:quant_compare_supply}.   
Specifically, we observe a large number of $180^\circ$ flips, because bounding box depth does not encode the front-facing direction of the object. Therefore, we evaluate a \textit{relaxed} angular error which does not penalize $180^\circ$ flips in the generated objects, results tabulated in~\cref{tab:quant_compare_supply}. We observe that 3D layout control baselines, LooseControl and Build-A-Scene show slightly improved results compared to \textit{LaRender}~\cite{zhan2025larender} and \textit{VODiff}~\cite{liang2025vodiff}, which are not orientation aware. \textit{Compass Control}~\cite{parihar2025compass} encodes orientation value through an adapter, hence performs better than the other baselines on angular error. In contrast, our OSCR representation explicitly encodes orientation in the image space using color-coding, thus enabling precise orientation control, performing favorably compared to all existing methods.

\begin{table}
\centering
\resizebox{\columnwidth}{!}{%
\begin{tabular}{@{}cccccc>{\columncolor{violet!20}}c@{}}
\toprule
\textbf{Baselines} & depth ord.$\uparrow$ & obj. score$\uparrow$  & angular err.$\downarrow$ & text align.$\uparrow$ & KID$(\times10^{-3})\downarrow$ & $180^\circ$ flip ang. err.$\downarrow$ \ \\ \midrule 
VODiff~\cite{liang2025vodiff} & $0.68$ & {19.70}  & $92.73$  & $29.51$ & $15.40$ & $41.38$       \\
LooseControl~\cite{loosecontrol}  & $0.82$ & $20.02$  & $89.88$  & $28.43$ & $14.32$ & $37.48$      \\
Build-A-Scene~\cite{eldesokey2024build} & $0.89$ & $21.0$  & $91.62$  & $28.05$ & $20.12$ & $32.23$        \\
LaRender~\cite{zhan2025larender} & $1.02$ & $21.83$ & $89.63$ & $30.20$ & $13.46$ & $41.19$       \\
\rowcolor{yellow!50} CompassControl~\cite{parihar2025compass} & $0.87$ & $20.60$ & $66.29$ & $29.76$ & $13.01$ & $35.79$ \\
\textbf{Ours} & $\mathbf{1.46}$ & $\mathbf{22.86}$  & $\mathbf{47.92}$  & $\mathbf{31.87}$ & $\mathbf{5.43}$ & $\mathbf{25.72}$   \\ 

\bottomrule
\end{tabular}%
}
\vspace{-1mm}
\caption{\textbf{Quantitative comparison:} In the main paper, we did not compare against methods that only enable partial 3D control:~\textit{Compass Control}~\cite{parihar2025compass} and \textit{ORIGEN}~\cite{min2025origen}. These baselines do not allow for 3D layout control, and primarily focus on object orientation control. While Compass Control allows for 2D layout control, ORIGEN does not, and hence it is not compatible with our quantitative evaluation (see \textbf{Evaluation metrics} in the main paper). Results for Compass Control are presented in yellow. It implicitly encodes orientation using an adapter, hence performs better than the other baselines in angular error. Further, we evaluate a \textit{relaxed} angular error, which does not penalize $180^\circ$ flips in the generated object (violet column). This caters to layout depth based methods~\textit{LooseControl}~\cite{loosecontrol} and \textit{Build-A-Scene}~\cite{eldesokey2024build}, which do not encode a front-facing direction for the objects, thus result in such $180^\circ$ flips. Our OSCR representation explicitly encodes orientation in the image space using color-coding, thus enabling precise orientation control, outperforming all baselines.}
\label{tab:quant_compare_supply}
\vspace{-4.0mm} 
\end{table}

\label{sec:ang_err}
\VA{
Aims of this section
\begin{itemize}
\item justify why LC and BAS have almost the same angular error as LaRender and VODiff. 
\item include the 180 degree flip agnostic results, where LC and BAS do better, but still worse than us.  
\end{itemize}
}

\begin{figure}[t]
  \centering
   \includegraphics[width=1.0\linewidth]{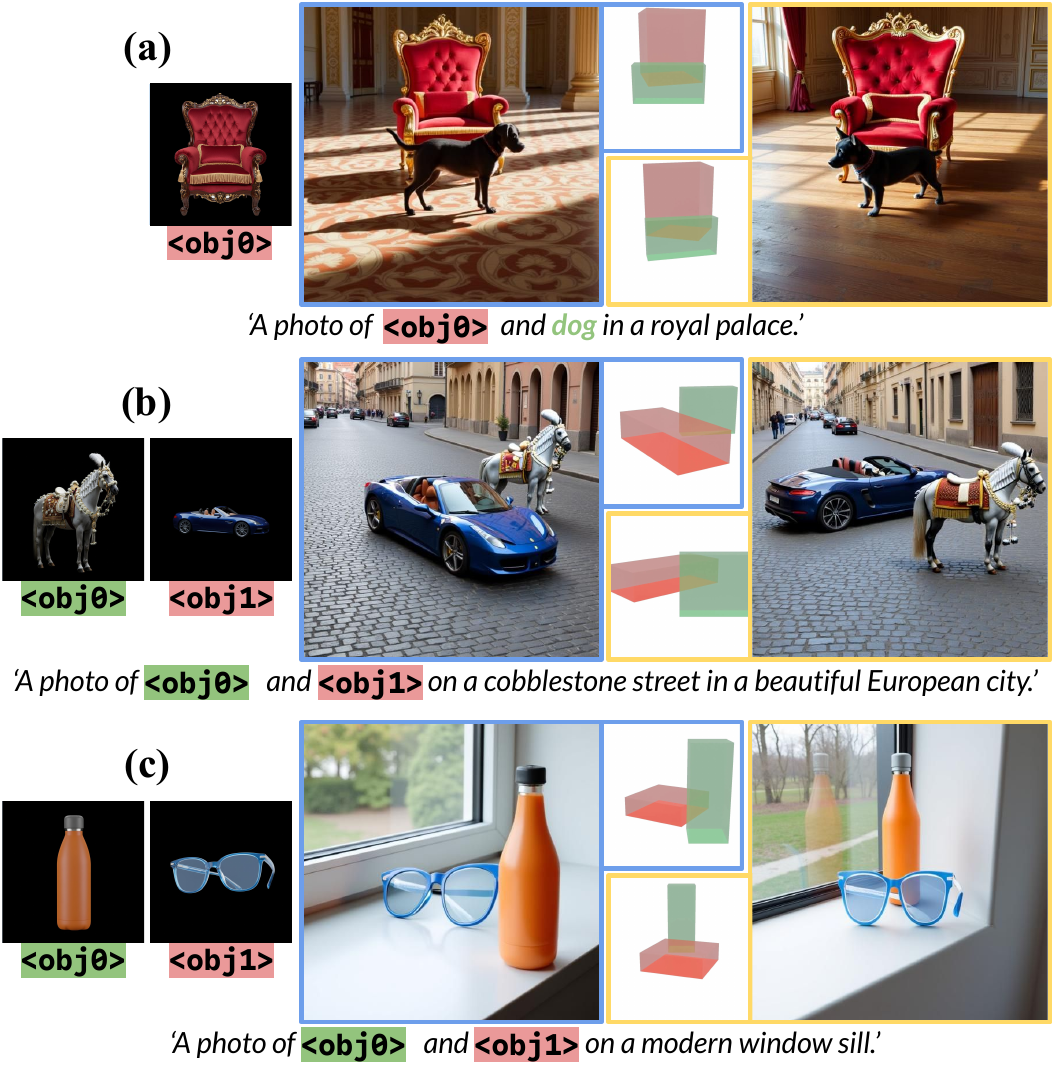}
   \caption{\textbf{Personalization:} We show that SeeThrough3D can be finetuned for occlusion-aware 3D control of personalized objects. This is achieved by learning a separate `subject' LoRA to fuse appearance attributes from personalized object image into the generation process, building upon prior work on conditioning diffusion transformers~\cite{tan2024ominicontrol,tan2025ominicontrol2,zhang2025easycontrol}. This approach achieves such personalized 3D control \textbf{without need for any test-time tuning}. As shown in (a), we can \textbf{compose objects from multiple modalities}, such as dog (text) and royal chair (image). Interestingly, our model can \textbf{personalize object categories not seen during} training, such as bottle and glasses in (c), indicating strong generalization.}
   \label{fig:personalization_supply}
\end{figure}

\begin{figure}[t]
  \centering
   \includegraphics[width=1.0\linewidth]{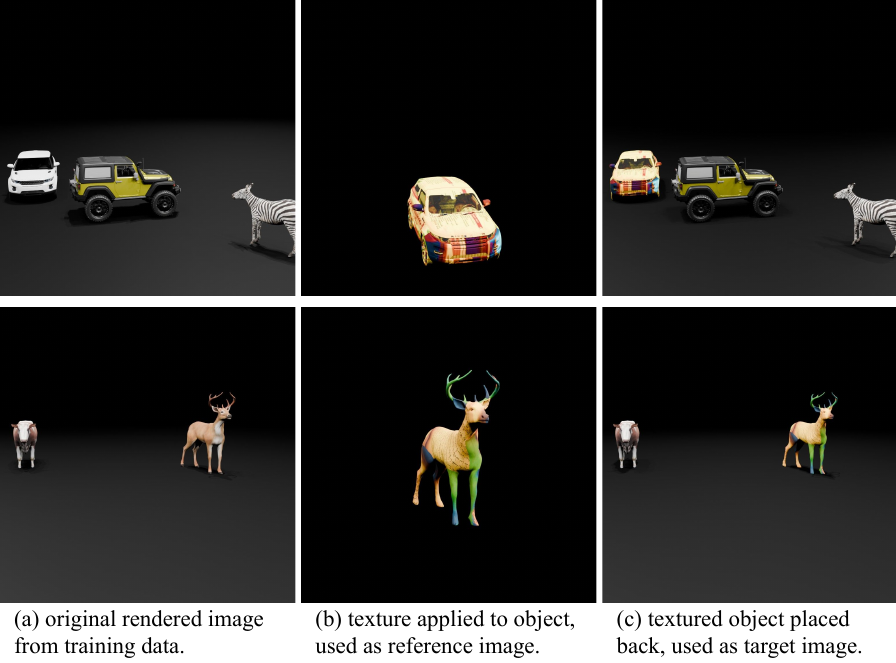}
   \caption{\textbf{Adapting the training dataset for personalization:} (a) given a rendered image from the training dataset, we randomly choose an object and apply a texture to it in Blender~\cite{blender} (see~\cref{fig:textures} for examples of generated textures). (b) We separately render the textured 3D asset, and use it as reference object condition. (c) Finally, the textured object is placed back into the original image, and used as ground truth target for training the model.} 
   \label{fig:personalization_dataset}
\end{figure}

\begin{figure}[t]
  \centering
   \includegraphics[width=1.0\linewidth]{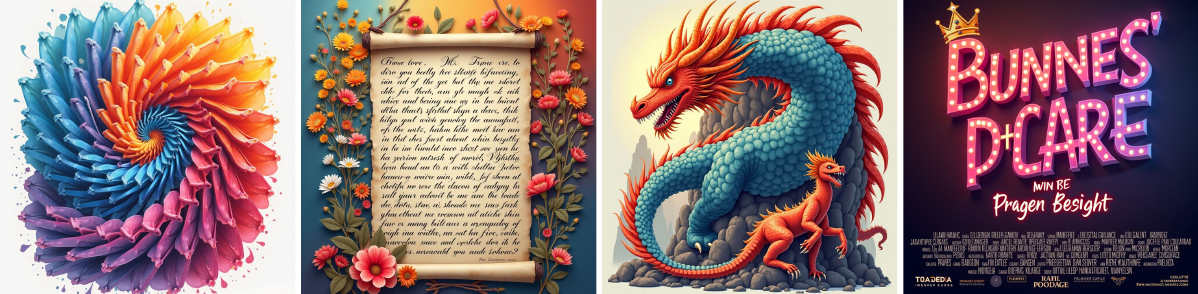}
   \caption{\textbf{Examples of generated textures:} We generate textures using FLUX~\cite{labs2025flux}, for preparing data for the personalization task. Notice how the textures contain high frequency features, induced by text and sharp patterns.} 
   \label{fig:textures}
\end{figure}

\begin{figure}[h]
  \centering
   \includegraphics[width=0.8\linewidth]{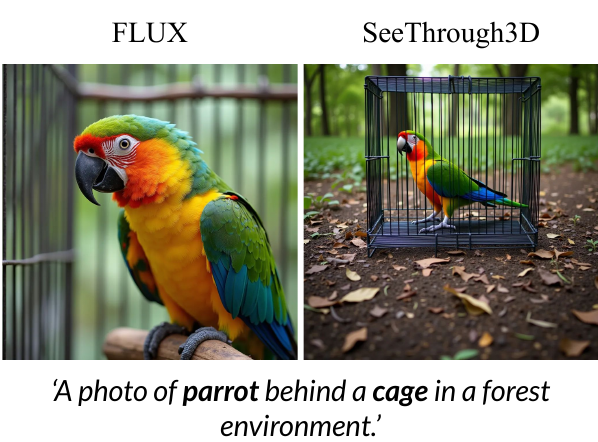}
   \caption{\textbf{Limitations:} (a) Our model is built upon FLUX~\cite{labs2025flux} which fails to generate some out of distribution cases, such as a parrot outside a cage. Consequently, our model also struggles to generate such cases.} 
   \label{fig:limitations}
\end{figure}

\section{Personalization}
\label{sec:personalization_supplement}
\VA{
Aims of this section
\begin{itemize}
\item show more personalization results. 
\item talk about how it generalizes to novel categories, not seen during training. 
\end{itemize}
}
We show that SeeThrough3D can be finetuned for occlusion-aware 3D control of personalized objects (see~\cref{fig:personalization_supply}). This is achieved by learning a separate `subject' LoRA to fuse appearance attributes from personalized object image into the generation process, building upon prior work on conditioning diffusion transformers~\cite{zhang2025easycontrol}. This approach achieves personalized 3D control \textbf{without need for any test-time tuning}. As shown in (a), we can \textbf{compose objects from multiple modalities}, such as dog (text) and royal chair (image). Interestingly, our model can \textbf{personalize object categories not seen during} training, such as bottle and glasses in (c), indicating strong generalization.

\begin{figure*}[t]
  \centering
   \includegraphics[width=1.0\linewidth]{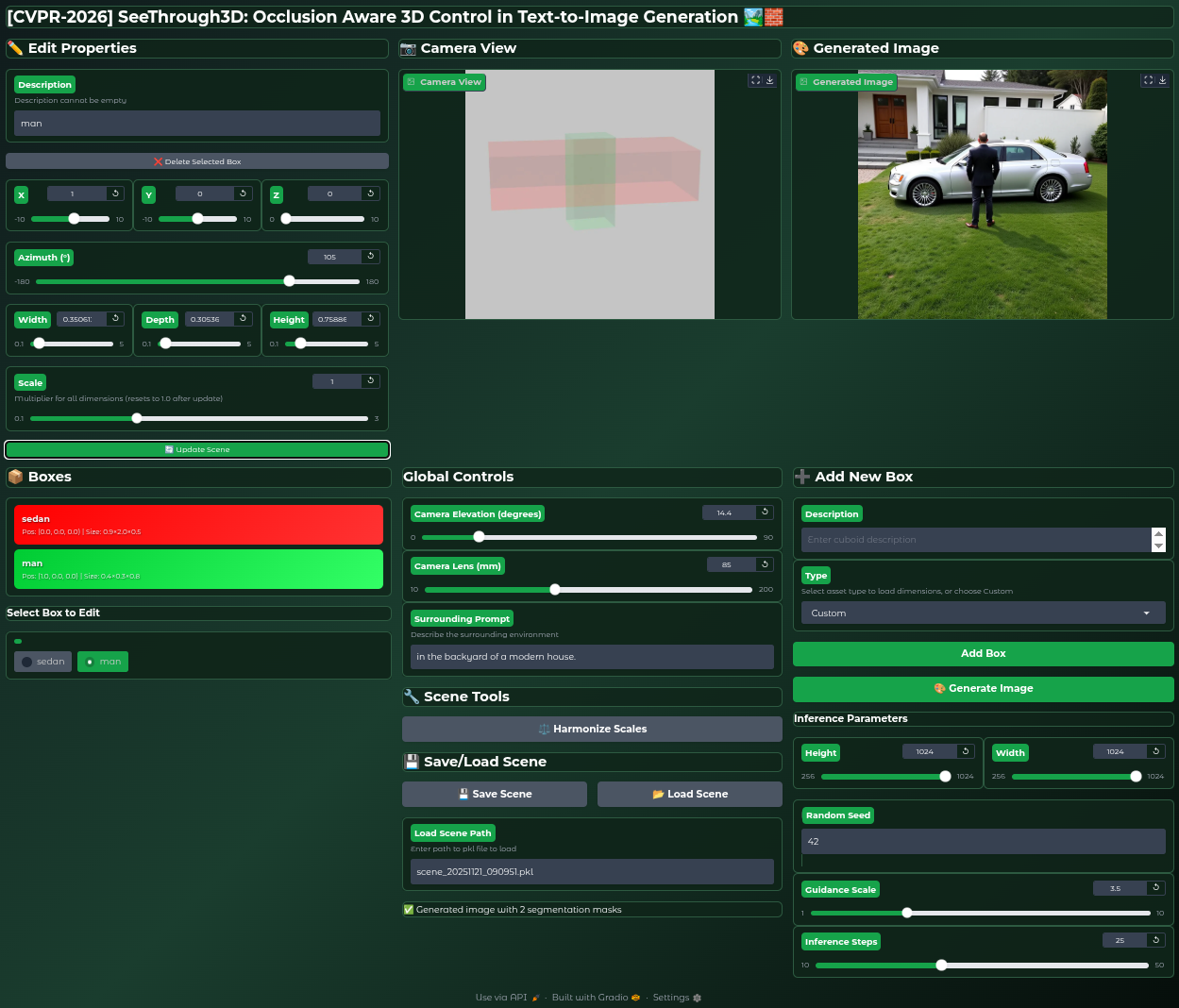}
   \caption{We built an intuitive UI to enable the user to easily design layouts for using our model. The UI enables the user to add objects, edit their placement, orientation and dimensions, and provide a text description for each object. Additionally, it allows the user to set the camera parameters.} 
   \label{fig:ui}
\end{figure*}

To train the personalization model, we suitably adapt our dataset for this task. Given a rendered image, we randomly choose an object and apply a texture to it in Blender~\cite{blender} (see~\cref{fig:personalization_dataset}(a,b)). For this, we generate a small set of textures using FLUX~\cite{labs2025flux} by prompting it to ensure high frequency details such as text and sharp patterns, some samples are shown in~\cref{fig:textures}. We separately render the textured 3D asset, and use it as the reference image condition (see~\cref{fig:personalization_dataset}(b)). The orientation of the reference object is slightly altered to enable the model to reason about its 3D placement, and not just copy pixels from reference image. Finally, the textured object is placed back into the original image, and used as ground truth target for training the model (see~\cref{fig:personalization_dataset}(c)), conditioned on the reference image.

\section{Additional qualitative comparisons}
\begin{figure*}[t]
  \centering
   \includegraphics[width=1.00\linewidth]{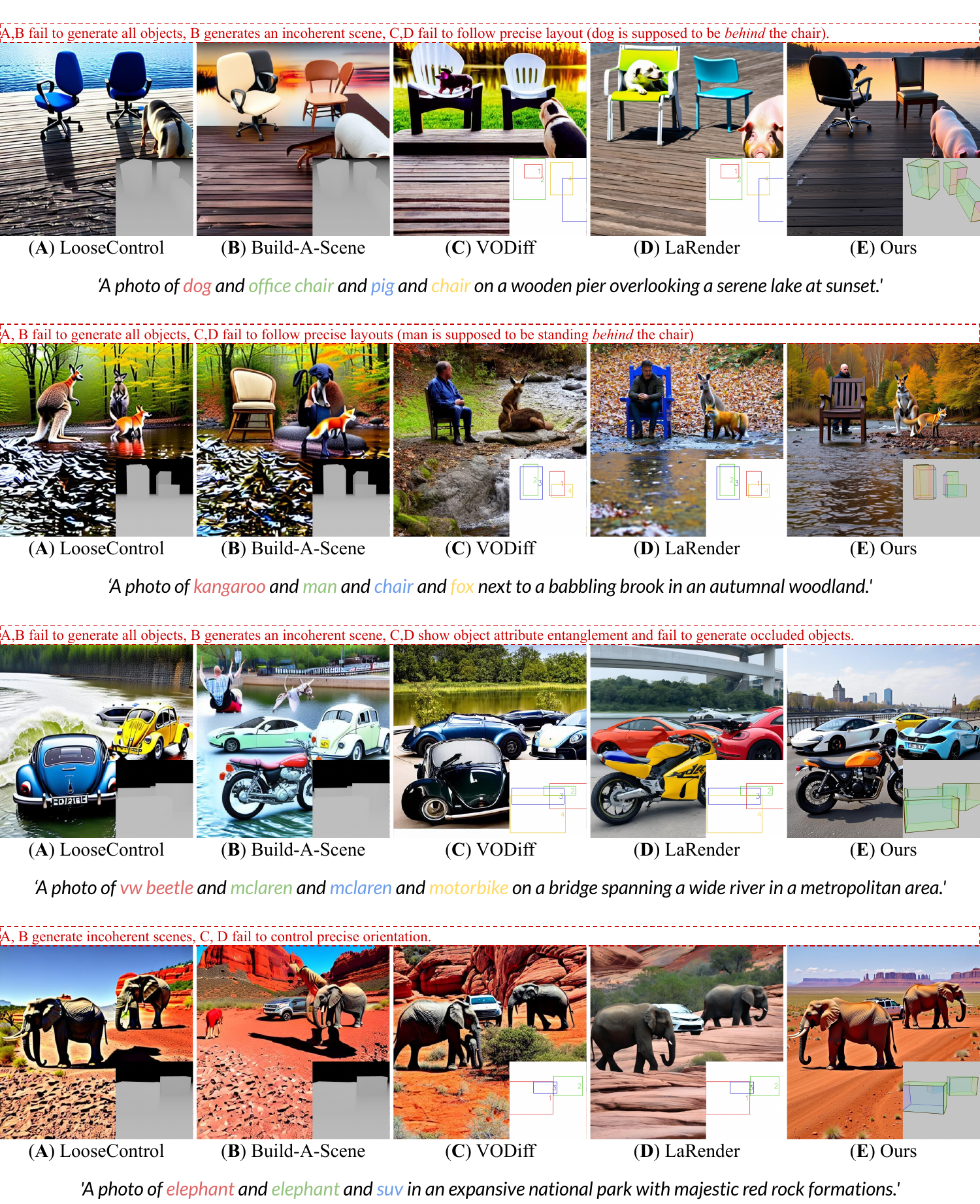}
   \caption{We present qualitative comparisons with the baselines from the main paper, which are categorized into \textbf{3D layout control:} \textit{LooseControl}~\cite{loosecontrol} and Build-A-Scene~\cite{eldesokey2024build}; \textbf{Occlusion control:} \textit{VODiff}~\cite{liang2025vodiff} and LaRender~\cite{zhan2025larender}. Each example has been analyzed with reference to layout adherence, occlusion consistency and realism (\textbf{red text} above each example).} 
   \label{fig:bulk_qualitative_comparisons_a}
\end{figure*}

\begin{figure*}[t]
  \centering
   \includegraphics[width=1.00\linewidth]{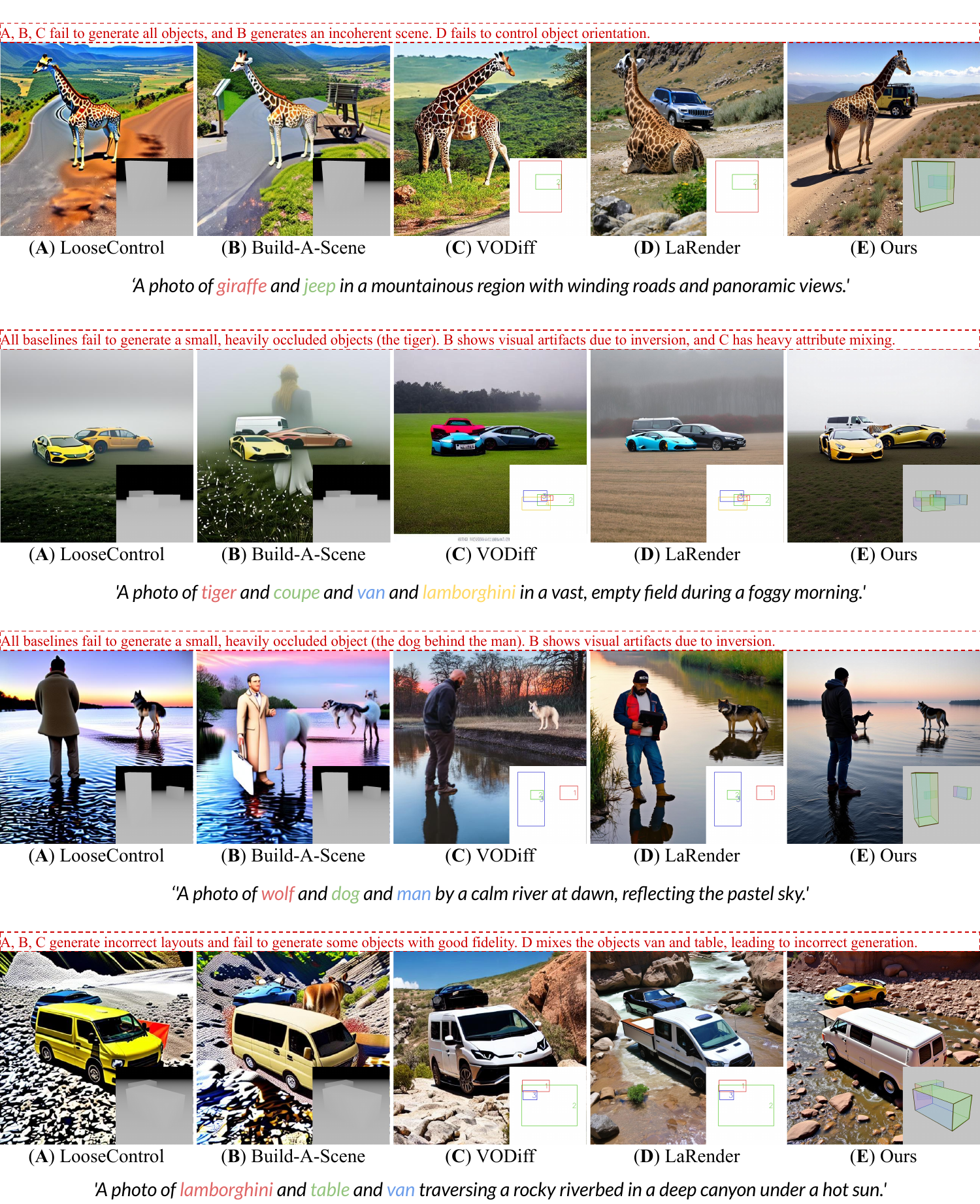}
   \caption{We present qualitative comparisons with the baselines from the main paper, which are categorized into \textbf{3D layout control:} \textit{LooseControl}~\cite{loosecontrol} and Build-A-Scene~\cite{eldesokey2024build}; \textbf{Occlusion control:} \textit{VODiff}~\cite{liang2025vodiff} and LaRender~\cite{zhan2025larender}. Each example has been analyzed with reference to layout adherence, occlusion consistency and realism (\textbf{red text} above each example).} 
   \label{fig:bulk_qualitative_comparisons_b}
\end{figure*}

We present additional qualitative comparisons with the main baselines in~\cref{fig:bulk_qualitative_comparisons_a,fig:bulk_qualitative_comparisons_b}. Each example has been analyzed with reference to layout adherence and occlusion consistency (red text above each example). Results indicate that SeeThrough3D generates realistic images following precise 3D layouts while maintaining occlusion consistency, and outperforms all baselines.

\section{Additional qualitative results}
\begin{figure*}[h]
  \centering
   \includegraphics[width=1.0\linewidth]{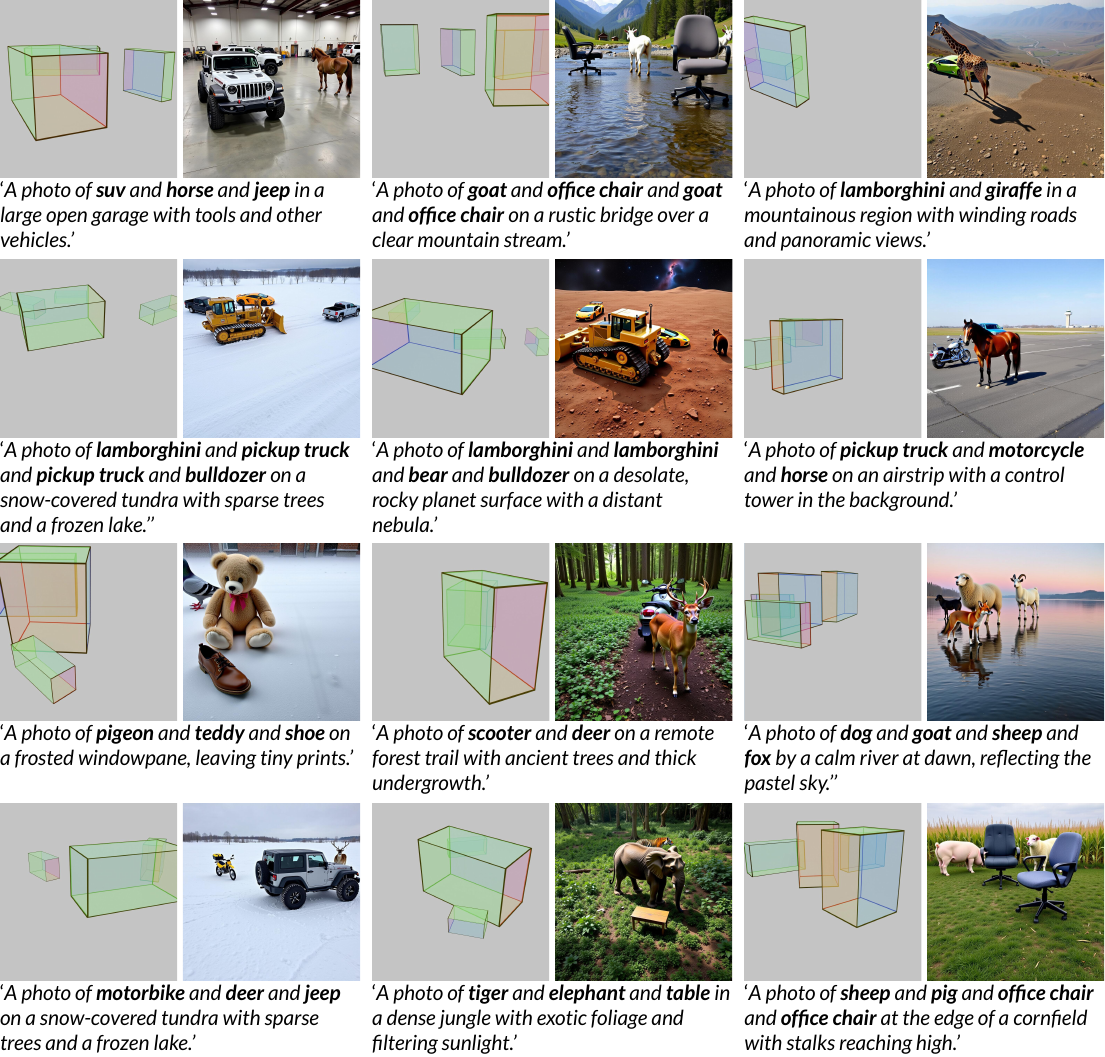}
   \caption{\textbf{Qualitative results:} We present additional results of our method. For each example, we have shown the OSCR layout alongside the generated image; the correspondence from boxes to individual objects has been omitted here for clarity. } 
   \label{fig:bulk_qualitative_results}
\end{figure*}

We present additional results of our method in~\cref{fig:bulk_qualitative_results}. For each example, we have shown the OSCR layout alongside the generated image; the correspondence from boxes to individual objects has been omitted here for clarity. 

\VA{
Aims of this section
\begin{itemize}
\item answer to if someone says it is incremental work -- we must highlight that with qualitative results and comparisons. 
\item implicitly, the large volume of examples should show that we have not done much cherry picking. 
\end{itemize} 
}

\section{User interface}
One of the motives of SeeThrough3D is to enable creative artists to precisely control various 3D elements of a generated image, such as scene layout and camera viewpoint. To ease the design process, we built an intuitive web interface, which allows the user to construct 3D layouts and control camera viewpoint. The interface allows the user to add boxes for various objects, edit their dimensions, 3D placement, and specify a text description for each object. The interface also comes with pre-defined template dimensions of common objects such cars, animals, etc. which can be used. A screenshot of the interface can be seen in~\cref{fig:ui}.



\section{Limitations}
\VA{
Aims of this section
\begin{itemize}
\item if FLUX messes up, then we mess up. this would also indicate that we build upon the strong prior of text-to-image models. show the parrot outisde cage examples.  
\item might fail to discertain top/behind in some cases of high camera elevation. 
\item fails to preserve consistency under layout changes / camera changes. 
\end{itemize} 
}
Since our method conditions a pretrained text-to-image model (FLUX~\cite{labs2025flux}), it is limited by the capabilities of the base model. For instance, FLUX sometimes fails to generate out of distribution cases, such as a parrot behind a bird-cage, with realistic occlusion. Consequently, our method, which is built upon the prior of FLUX, also fails to generate such cases, as shown in~\cref{fig:limitations}. Additionally, personalization requires that all the reference image tokens be present in the transformer's context; this leads to higher VRAM requirements, especially for multi-subject personalization.